\newcount\Comments
\documentclass[11pt]{article} 
\usepackage{tcolorbox}
\usepackage{bbm}
\usepackage{amsmath}
\usepackage{amssymb}
\usepackage{algorithm}
\usepackage{mathtools}
\usepackage{amsthm}
\usepackage{comment}
\usepackage{natbib}
\usepackage{authblk}

\PassOptionsToPackage{algo2e,ruled, linesnumbered}{algorithm2e}
\RequirePackage{algorithm2e}

\usepackage[letterpaper,top=2cm,bottom=2cm,left=3cm,right=3cm,marginparwidth=1.75cm]{geometry}

\usepackage[colorlinks=true, allcolors=blue]{hyperref}

\DeclarePairedDelimiter\ceil{\lceil}{\rceil}

\newcommand{\reals}{\mathbb{R}}
\newcommand{\naturals}{\mathbb{N}}

\usepackage{color}
\definecolor{darkgreen}{rgb}{0,0.5,0}
\definecolor{purple}{rgb}{1,0,1}
\newcommand{\kibitz}[2]{\ifnum\Comments=1\textcolor{#1}{#2}\fi}

\newcommand{\Acal}{\mathcal{A}}

\newcommand{\Hcal}{\mathcal{H}}

\newcommand{\Zcal}{\mathcal{Z}}
\newcommand{\Xcal}{\mathcal{X}}
\newcommand{\Ycal}{\mathcal{Y}}

\DeclareMathOperator*{\argmin}{arg\,min}

\DeclareMathOperator*{\expect}{\operatorname{\mathbb{E}}}

\newcommand{\indicator}{\mathbbm{1}}

\newtheorem{theorem}{Theorem}

\newtheorem{definition}{Definition}
\newtheorem{lemma}[theorem]{Lemma}

\newtheorem*{theorem*}{Theorem}
\newtheorem*{remark*}{Remark}
\usepackage{natbib}

\title{A Combinatorial Characterization of Supervised Online Learnability}

\author[ ]{Vinod Raman, Unique Subedi, Ambuj Tewari}
\affil[ ]{Department of Statistics, University of Michigan}
\affil[ ]{\texttt{\{vkraman, subedi, tewaria\}@umich.edu}}
\date{}

\begin{document}

\maketitle

\begin{abstract}
We study the online learnability of hypothesis classes with respect to arbitrary, but bounded loss functions. No characterization of online learnability is known at this level of generality. We give a new scale-sensitive combinatorial dimension, named the sequential minimax dimension, and show that it gives a tight quantitative characterization of online learnability. In addition, we show that the sequential minimax dimension subsumes most existing combinatorial dimensions in online learning theory.
\end{abstract}

\section{Introduction}
In the supervised online learning model, a learner plays a repeated game against an adversary over $T \in \mathbb{N}$ rounds. In each round $t \in [T]$, an adversary picks a labeled example $(x_t, y_t) \in \mathcal{X} \times \mathcal{Y}$ and reveals $x_t$ to the learner. The learner observes $x_t$, picks a probability measure $\mu_t$ over the prediction space $\mathcal{Z}$, and then makes a randomized prediction $z_t \sim \mu_t$. Finally, the adversary reveals the true label $y_t$ and the learner suffers the loss $\ell(y_t, z_t)$, where $\ell: 
\mathcal{Y} \times \mathcal{Z} \rightarrow \mathbb{R}_{\geq 0}$ is some pre-specified, bounded loss function. For a hypothesis class $\Hcal \subseteq \Zcal^{\Xcal}$ known apriori to the learner, the goal of the learner is to make predictions such that its expected regret, defined as the difference between the expected cumulative loss of the learner's predictions and that of the best-fixed hypothesis in $\Hcal$, is small. We say that a tuple $(\mathcal{X}, \mathcal{Y}, \mathcal{Z}, \Hcal, \ell)$ is \textit{online learnable} if there exists an online learner such that its expected regret is a sublinear function of $T$, for any strategy of the adversary.

In this work, we are interested in characterizing online learnability in full generality. The only known sequential complexity measure for an arbitrary learning problem $(\mathcal{X}, \mathcal{Y}, \mathcal{Z}, \Hcal, \ell)$ is the sequential Rademacher complexity of the loss class $\ell \circ \Hcal := \{(x, y) \mapsto \ell(y, h(x)): h \in \mathcal{H}\}$.  In particular, \cite{rakhlin2015online} show that if the sequential Rademacher complexity of the loss class $\ell \circ \mathcal{H}$ is a sublinear function of $T$, then $(\mathcal{X}, \mathcal{Y}, \mathcal{Z}, \Hcal, \ell)$ is online learnable. However, even for natural problems like online multiclass classification \citep{hanneke2023multiclass} and linear regression \citep{raman2024operator}, sublinear sequential Rademacher complexity is not \textit{necessary} for online learnability. Accordingly, we investigate the following question.
\begin{center}
What are necessary and sufficient conditions for $(\mathcal{X}, \mathcal{Y}, \mathcal{Z}, \Hcal, \ell)$ to be online learnable?
\end{center}
In many settings, online learnability is characterized in terms of combinatorial dimensions. For instance, when $\mathcal{Z} = \mathcal{Y}$ and $\ell(y, z) = \mathbbm{1}\{y \neq z\}$, online learnability is characterized in terms of Littlestone dimension of $\mathcal{H} \subseteq \mathcal{Z}^{\mathcal{X}}$, henceforth denoted as $\texttt{L}(\mathcal{H})$. That is,  $\mathcal{H} \subseteq \Zcal^{\Xcal}$ is online learnable if and only if $\texttt{L}(\mathcal{H}) < \infty$ \citep{Littlestone1987LearningQW, DanielyERMprinciple, hanneke2023multiclass}. Similarly, when $\mathcal{Z} = \mathcal{Y} =[-1,1]$ and $\ell(y,z) = |y-z|$, the \textit{sequential} fat-shattering dimension of  $\mathcal{H} \subseteq \mathcal{Z}^{\mathcal{X}}$, denoted $\texttt{sfat}_{\gamma}(\mathcal{H})$, characterizes the online learnability of $\mathcal{H}$.
That is,   $\mathcal{H} \subseteq \mathcal{Z}^{\mathcal{X}}$ is online learnable if and only if $\texttt{sfat}_{\gamma}(\mathcal{H}) < \infty$ at every scale $\gamma > 0$ \citep{rakhlin2015online}. 

Remarkably, the Littlestone and sequential fat-shattering dimensions also provide a tight \textit{quantitative} characterization of online learnability, appearing in both the lower and upperbounds of their respective minimax expected regret. This motivates the next question we investigate. 
\begin{center}
Given any tuple $(\mathcal{X}, \mathcal{Y}, \Zcal, \mathcal{H}, \ell)$, is there a combinatorial dimension that provides a tight quantitative characterization of the online learnability of $\mathcal{H}$? 
\end{center}
\noindent Guided by these two questions, we make the following contributions in this paper. 
\begin{itemize}
    \item[(i)] We give a new scale-sensitive dimension named the sequential minimax dimension (SMdim) and show that it provides a tight quantitative characterization of online learnability of any tuple  $(\mathcal{X}, \mathcal{Y}, \mathcal{Z}, \Hcal, \ell)$. 
    \item[(ii)] Using the SMdim, we construct a generic online learner for any tuple $(\mathcal{X}, \mathcal{Y}, \mathcal{Z}, \mathcal{H}, \ell)$.
    \item[(iii)] We show that existing combinatorial dimensions are special instances of the SMdim. This includes the case where $\mathcal{Z} = \mathcal{Y}$, like the Littlestone and sequential fat-shattering dimensions, as well as the case where $\mathcal{Z} \neq \mathcal{Y}$, like the $(k+1)$-Littlestone dimension from \cite{moran2023list} and Measure shattering dimension from \cite{raman2023online}.  
\end{itemize}

To prove the sufficiency of the SMdim, we combine and extend the algorithmic ideas from \cite{DanielyERMprinciple} and \cite{hanneke2023multiclass} to construct a generic online learner.  To prove the necessity of the SMdim, we construct a hard stream given any online learner. Finally, we use combinatorial arguments and results from discrete geometry to show that the SMdim reduces to existing dimensions.

\subsection{Related Works}
Characterizing learnability in terms of complexity measures has a long rich history in statistical learning theory, originating from the seminal work of \cite{vapnik1971uniform}. In online learning, \cite{Littlestone1987LearningQW} showed that a combinatorial parameter, later named the Littlestone dimension, provides a quantitative characterization of online binary classification in the realizable setting. Twenty-two years later, \cite{ben2009agnostic} proved that the Littlestone dimension also provides a tight quantitative characterization of online binary classification in the agnostic setting. \cite{DanielyERMprinciple} generalized the Littlestone dimension to multiclass classification and showed that it fully characterizes online learnability when the label space is finite. Recently, \cite{hanneke2023multiclass} proved that the multiclass extension of the Littlestone dimension characterizes multiclass learnability under the $0$-$1$ loss even when the label space is unbounded. In a parallel line of work, \cite{rakhlin2015online, rakhlin2015sequential} defined the sequential fat-shattering dimension and showed that it tightly characterizes the online learnability of scalar-valued regression with respect to the absolute value loss. In addition, they defined a general complexity measure called the sequential Rademacher complexity and proved that it upperbounds the minimax expected regret of any supervised online learning game. In a similar spirit, we define a \emph{combinatorial dimension} that upper and lowerbounds the minimax expected regret of any supervised online learning game. 

The proof techniques in online learning are generally constructive and result in beautiful algorithms such as Follow The (Regularized) Leader, Hedge, Multiplicative Weights, Online Gradient Descent, and so forth. In online binary classification, \cite{Littlestone1987LearningQW} proposed the Standard Optimal Algorithm and proved its optimality in the realizable setting. \cite{DanielyERMprinciple} and \cite{rakhlin2015online} generalize this algorithm to multiclass classification and scalar-valued regression respectively. The ideas of the Standard Optimal Algorithm is foundational in online learning and still appears in more recent works by \cite{moran2023list}, \cite{ filmus2023optimal}, and \cite{raman2023online}. A common theme in these variants of the Standard Optimal Algorithm is their use of combinatorial dimensions to make predictions. Similarly, \cite{rakhlin_relax_n_randomize} use the sequential Rademacher complexity to directly construct a generic online learner in the agnostic setting. However, their online learner requires the sequential Rademacher complexity of the loss class to be sublinear in $T$, and thus does not work for arbitrary tuples $(\Xcal, \Ycal, \Zcal, \Hcal, \ell)$. Closing this gap, we define a new scale-sensitive dimension, named the sequential minimax dimension, and use it to give a generic online learner for any tuple $(\Xcal, \Ycal, \Zcal, \Hcal, \ell)$. 

\section{Preliminaries}
\subsection{Notation}
Let $\Xcal$ denote the instance space, $\Ycal$ denote the label space, and $\Zcal$ denote the prediction space. For a sigma algebra $\sigma(\Zcal)$ on the prediction space $\Zcal$, define $\Pi(\Zcal)$ to be the set of all distributions on $(\Zcal, \sigma(\Zcal))$. For any set $S \in \sigma(\mathcal{Z})$, let $S^c$ denote its complement.    Let $\mathcal{H} \subseteq \mathcal{Z}^{\mathcal{X}}$ denote an arbitrary hypothesis class consisting of predictors $h: \mathcal{X} \rightarrow \mathcal{Z}$ that maps an instance to a prediction. Given any prediction $z \in \Zcal$ and a label $y \in \Ycal$, we consider a loss function $\ell : \Ycal \times \Zcal \to \reals_{\geq 0}$.  We put no restrictions on the loss function $\ell$, except that it is bounded, $\sup_{y, z} \ell(y, z) \leq c$ for some $c \in \reals_{>0}$. In particular, the loss can asymmetric, and therefore we reserve the first argument for the label and the second argument for the prediction. Finally,  $[N] :=\{1, 2, \ldots, N\}$.

\subsection{Supervised Online Learning}
In the supervised online learning setting, an adversary plays a sequential game with the learner over $T$ rounds. In each round $t \in [T]$, the adversary selects a labeled instance $(x_t, y_t) \in \mathcal{X} \times \Ycal$ and reveals $x_t$ to the learner. The learner picks a probability measure $\mu_t \in \Pi(\mathcal{Z})$ and then makes a randomized prediction $z_t \sim \mu_t$. Finally, the adversary reveals the feedback $y_t$, and the learner suffers the loss $\ell(y_t, z_t)$. Given a hypothesis class  $\mathcal{H} \subseteq \mathcal{Z}^{\mathcal{X}}$, the goal of the learner is to output randomized predictions $z_t$ such that its expected cumulative loss is close to the smallest possible cumulative loss over hypotheses in $\mathcal{H}$. 

To formally define an online learning algorithm, we follow the convention of  \cite[Chapter 4]{cesa2006prediction}. 

\begin{definition}[Supervised Online Learning Algorithm]
 \noindent A supervised online learning algorithm is a deterministic mapping $\Acal : (\Xcal \times \Ycal )^{\star} \times \Xcal \to \Pi(\Zcal)$ that maps past examples  and the newly revealed instance $x \in \Xcal$ to a probability measure $\mu \in \Pi(\Zcal)$. The learner then randomly samples $z \sim \mu$ to make a prediction.
\end{definition} 
\noindent Although $\Acal$ is a deterministic mapping, the prediction $z \sim \mu$ is random. Restricting the range of $\Acal$ to be the set of Dirac measures on $\Zcal$ yields a deterministic online learner. When the context is clear, with a slight abuse of notation, we use $\Acal(x)$ to denote the random sample $z$ drawn from the distribution that $\Acal$ outputs. We say that $\mathcal{H}$ is online learnable with respect to $\ell$ if there exists an online learning algorithm $\mathcal{A}$ with ``small" \textit{expected regret}:

$$\texttt{R}_{\mathcal{A}}(T,\mathcal{H\, }, \ell) := \sup_{(x_1, y_1), \ldots, (x_T, y_T)} \left( \sum_{t=1}^T \mathbb{E}\big[ \ell(y_t, \Acal(x_t)) \big]- \inf_{h \in \mathcal{H}}\sum_{t=1}^T  \ell(y_t, h(x_t)) \right).$$

\begin{definition}[Supervised Online Learnability]\label{def:agnOL}
\noindent A hypothesis class $\Hcal \subseteq \Zcal^{\Xcal}$ is online learnable if and only if  $\inf_{\mathcal{A}} \emph{\texttt{R}}_{\mathcal{A}}(T, \mathcal{H}, \ell) = o(T).$
\end{definition}

\subsection{Combinatorial dimensions}
In online learning theory, combinatorial dimensions play an important role in providing crisp quantitative characterizations of learnability. Formally, we define a combinatorial dimension as a function $\texttt{D}$ that maps $(\Hcal, \ell)$ to $\naturals \cup \{0,\infty\}$ and satisfies the following two properties: (1) $\mathcal{H}$ is online learnable with respect to $\ell$ if and only if $\texttt{D}(\mathcal{H}, \ell) < \infty$ and (2) the minimax expected regret $\inf_{\Acal} \texttt{R}_{\Acal}( T, \Hcal, \ell)$ depends only on $\texttt{D}(\Hcal, \ell)$ and $T$. In particular, $\inf_{\Acal} \texttt{R}_{\Acal}( T, \Hcal, \ell)$ should not depend on any other property of the tuple $(\Xcal, \Ycal, \Zcal, \Hcal, \ell)$ such as $|\mathcal{Y}|$ or $|\Zcal|$. We also allow a combinatorial dimension to take a scale parameter as an input. That is, a scale-sensitive combinatorial dimension is a function $\texttt{D}$ that maps $(\Hcal, \ell)$ and a scale  $\gamma>0 $ to $\naturals \cup \{0,\infty\}$  with the following two properties: (1) $\mathcal{H}$ is online learnable with respect to $\ell$ if and only if $\texttt{D}(\mathcal{H}, \ell, \gamma) < \infty$ for every $\gamma > 0$ and (2) the minimax expected regret $\inf_{\Acal} \texttt{R}_{\Acal}( T, \Hcal, \ell)$ can be lower- and upperbounded in terms of $T$ and $\texttt{D}(\mathcal{H}, \ell, \cdot)$. Our definition of a combinatorial dimension is similar to the definition in \citep{ben2019learnability} with two key differences. In particular, the notion of dimension in \citep{ben2019learnability} requires $\texttt{D}(\Hcal, \ell)$ to satisfy the finite-character property (see Section \ref{sec:finitechar}), but does not require it to provide a quantitative characterization of learnability.

Nevertheless, our definition of dimension also captures all existing combinatorial dimensions in online learning theory, such as the Littlestone and sequential fat-shattering dimension. These dimensions are typically defined in terms of trees, a basic combinatorial object that captures the temporal dependence inherent in online learning. Given an instance space $\mathcal{X}$ and a (potentially uncountable) set of objects $\mathcal{M}$, a $\mathcal{X}$-valued, $\mathcal{M}$-ary tree $\mathcal{T}$ of depth $T$ is a complete rooted tree such that (1) each internal node is labeled by an instance $x \in \mathcal{X}$ and (2) for every internal node and object $m \in \mathcal{M}$, there is an outgoing edge indexed by $m$. Such a tree can be identified by a sequence $(\mathcal{T}_1, ..., \mathcal{T}_T)$ of labeling functions $\mathcal{T}_t:\mathcal{M}^{t-1} \rightarrow \mathcal{X}$ which provide the labels for each internal node. A path of length $T$ is given by a sequence of objects $m = (m_1,..., m_T) \in \mathcal{M}^T$. Then, $\mathcal{T}_t(m_1, ..., m_{t-1})$ gives the label of the node by following the path $(m_1, ..., m_{t-1})$ starting from the root node, going down the edges indexed by the $m_t$'s.  We let $\mathcal{T}_1 \in \mathcal{X}$ denote the instance labeling the root node. For brevity, we define $m_{<t} = (m_1, ..., m_{t-1})$ and therefore write $\mathcal{T}_t(m_1, ..., m_{t-1}) = \mathcal{T}_t(m_{<t})$. Analogously, we let $m_{\leq t} = (m_1, ..., m_{t})$.

Often, it is useful to label the edges of a tree with some \textit{auxiliary} information. Given a $\mathcal{X}$-valued, $\mathcal{M}$-ary tree $\mathcal{T}$ of depth $T$ and a (potentially uncountable) set of objects $\mathcal{N}$, we can formally label the edges of $\mathcal{T}$ using objects in $\mathcal{N}$ by considering a sequence $(f_1, ..., f_T)$ of edge-labeling functions  $f_t: \mathcal{M}^{t} \rightarrow \mathcal{N}$. For each depth $t \in [T]$, the function $f_t$ takes as input a path $m_{\leq t}$ of length $t$ and outputs an object in $\mathcal{N}$. Accordingly, we can think of the object $f_t(m_{\leq t})$ as labeling the edge indexed by $m_t$ after following the path $m_{< t}$ down the tree. We now use this notation to rigorously define existing combinatorial dimensions in online learning. 

We start with the Littlestone dimension, which is known to characterize binary/multiclass online classification. In this setting, we take $\mathcal{Y} = \mathcal{Z}$ and $\ell(y, z) = \mathbbm{1}\{y \neq z\}.$

\begin{definition} [Littlestone dimension \citep{Littlestone1987LearningQW, DanielyERMprinciple}]\label{def:ldim}
\noindent Let $\mathcal{T}$ be a complete, $\mathcal{X}$-valued, $\{\pm 1\}$-ary tree of depth $d$. The tree $\mathcal{T}$ is shattered by $\mathcal{H} \subseteq \Zcal^{\Xcal}$  if there exists a sequence $(f_1, ..., f_d)$ of edge-labeling functions  $f_t: \{\pm 1\}^{t} \rightarrow \mathcal{Y}$  such that for every path $\sigma = (\sigma_1, ..., \sigma_d) \in \{\pm 1\}^d$, there exists a hypothesis $h_{\sigma} \in \mathcal{H}$ such that for all $t \in [d]$,  $h_{\sigma}(\mathcal{T}_t(\sigma_{<t})) = f_t(\sigma_{\leq t})$ and $f_t((\sigma_{< t}, -1)) \neq f_t((\sigma_{< t}, +1))$. The Littlestone dimension of $\mathcal{H}$, denoted $\emph{\texttt{L}}(\mathcal{H})$, is the maximal depth of a tree $\mathcal{T}$ that is shattered by $\mathcal{H}$. If there exists shattered trees of arbitrarily large depth, we say $\emph{\texttt{L}}(\mathcal{H}) = \infty$.
\end{definition}

For online regression, where we take $\mathcal{Z} = \mathcal{Y} = [-1, 1]$ and $\ell(y, z) = |y - z|$, online learnability is characterized by the sequential-fat shattering (seq-fat) dimension.

\begin{definition} [Sequential fat-shattering dimension \citep{rakhlin2015online}]\label{def:sfat}
\noindent Let $\mathcal{T}$ be a complete, $\mathcal{X}$-valued, $\{\pm 1\}$-ary tree of depth $d$ and fix $\gamma \in (0, 1]$. The tree $\mathcal{T}$ is $\gamma$-shattered by $\mathcal{H} \subseteq \mathcal{Z}^{\Xcal}$  if there exists a sequence $(f_1, ..., f_d)$ of edge-labeling functions  $f_t: \{\pm 1\}^{t} \rightarrow \mathcal{Y}$  such that for every path $\sigma = (\sigma_1, ..., \sigma_d) \in \{\pm 1\}^d$, there exists a hypothesis $h_{\sigma} \in \mathcal{H}$ such that for all $t \in [d]$, $\sigma_t(h_{\sigma}(\mathcal{T}_t(\sigma_{<t}))  - f_t(\sigma_{\leq t})) \geq \gamma$ and $f_t((\sigma_{< t}, -1)) = f_t((\sigma_{< t}, +1))$. The sequential fat-shattering dimension of $\mathcal{H}$ at scale $\gamma$, denoted $\emph{\texttt{sfat}}_{\gamma}(\mathcal{H})$, is the maximal depth of a tree $\mathcal{T}$ that is $\gamma$-shattered by $\mathcal{H}$. If there exists $\gamma$-shattered trees of arbitrarily large depth, we say that $\emph{\texttt{sfat}}_{\gamma}(\mathcal{H}) = \infty$.
\end{definition}

 Recently, \cite{moran2023list} study list online classification, where we take $\mathcal{Z} = \{S: S\subset \mathcal{Y}, |S| \leq k\}$ and $\ell(y, z) = \mathbbm{1}\{y \notin z\}$. Here, they show that the $(k+1)$- Littlestone dimension, characterizes online learnability of a hypothesis class $\mathcal{H} \subseteq \mathcal{Z}^{\mathcal{X}}$. 

\begin{definition}[$(k+1)\text{-Littlestone dimension}$ \citep{moran2023list}]\label{def:kldim}
\noindent Let $\mathcal{T}$ be a complete, $\mathcal{X}$-valued, $[k+1]$-ary tree of depth $d$. The tree $\mathcal{T}$ is shattered by $\mathcal{H} \subseteq \mathcal{Z}^{\Xcal}$  if there exists a sequence $(f_1, ..., f_d)$ of edge-labeling functions  $f_t: [k+1]^{t} \rightarrow \mathcal{Y}$  such that for every path $p = (p_1, ..., p_d) \in [k+1]^d$, there exists a hypothesis $h_{p} \in \mathcal{H}$ such that for all $t \in [d]$, $f_t(p_{\leq t})) \in h_{\sigma}(\mathcal{T}_t(\sigma_{<t}))$  and for all distinct $i, j \in [k+1]$,  $f_t((p_{< t}, i)) \neq f_t((p_{< t}, j))$. The $(k+1)$-Littlestone  dimension of $\mathcal{H}$ denoted $\emph{\texttt{L}}_{k+1}(\mathcal{H})$, is the maximal depth of a tree $\mathcal{T}$ that is shattered by $\mathcal{H}$. If there exists shattered trees of arbitrarily large depth, we say that $\emph{\texttt{L}}_{k+1}(\mathcal{H}) = \infty$.
\end{definition}


Finally, in the ``flip" of list online classification, where $\mathcal{Y} \subset \sigma(\mathcal{Z})$ is some collection of measurable subsets of $\mathcal{Z}$ and $\ell(y, z) = \mathbbm{1}\{z \notin y\}$, \cite{raman2023online} show that the Measure shattering dimension characterizes online learnability of a hypothesis class $\mathcal{H} \subseteq \mathcal{Z}^{\mathcal{X}}$. 

\begin{definition} [Measure shattering dimension \citep{raman2023online}]
\noindent Let $\mathcal{T}$ be a complete $\mathcal{X}$-valued, $\Pi(\mathcal{Z})$-ary tree of depth $d$, and fix $\gamma \in (0,1] $. The tree $\mathcal{T}$ is $\gamma$-shattered by $\mathcal{H} \subseteq \Zcal^{\Xcal}$  if there exists a sequence $(f_1, ..., f_d)$ of edge-labeling set-valued functions  $f_t: \Pi(\mathcal{Z})^{t} \rightarrow \mathcal{Y}$  such that for every path $\mu = (\mu_1, ..., \mu_d) \in \Pi(\mathcal{Z})^d$, there exists a hypothesis $h_{\mu} \in \mathcal{H}$ such that for all $t \in [d]$,  $h_{\mu}(\mathcal{T}_t(\mu_{<t})) \in f_t(\mu_{\leq t})$ and $\mu_t(f_t(\mu_{\leq t})) \leq 1 - \gamma$. The Measure Shattering dimension of $\mathcal{H}$ at scale $\gamma$, denoted $\emph{\texttt{MS}}_{\gamma}(\mathcal{H}, \mathcal{Y})$, is the maximal depth of a tree $\mathcal{T}$ that is $\gamma$-shattered by $\mathcal{H}$. If there exists $\gamma$-shattered trees of arbitrarily large depth, we say $\emph{\texttt{MS}}_{\gamma}(\mathcal{H}, \mathcal{Y}) = \infty$. 
\end{definition}


\section{A Unifying Combinatorial Dimension} \label{sec:unify}
In this work, we are interested in finding a combinatorial dimension that characterizes the online learnability of an arbitrary tuple $(\Xcal, \Ycal, \Zcal, \Hcal, \ell)$. Inspired by the definition of minimax expected regret, we define a dimension, termed the sequential minimax dimension, that provides an adversary with a strategy against every possible move of the learner. Since the learner plays measures in $\Pi(\mathcal{Z})$, this amounts to defining a tree where each internal node has an outgoing edge labeled by an element of $\mathcal{Y}$ for every measure in $\Pi(\mathcal{Z})$. For any prediction $\mu \in \Pi(\mathcal{Z})$ by the learner, the label on the edge associated to $\mu$ gives the element $y \in \mathcal{Y}$ that the adversary should play to force the learner to suffer a large expected loss. Definition \ref{def:SMdim} formalizes this idea. 

\begin{definition}[Sequential minimax dimension]\label{def:SMdim}
\noindent Let $\mathcal{T}$ be a complete $\mathcal{X}$-valued, $\Pi(\mathcal{Z})$-ary tree of depth $d$, and fix $\gamma > 0$. The tree $\mathcal{T}$ is $\gamma$-shattered by $\mathcal{H} \subseteq \Zcal^{\Xcal}$ with respect to $\ell: \mathcal{Y} \times \mathcal{Z} \rightarrow \mathbb{R}_{\geq0}$ if there exists a sequence $(f_1, ..., f_d)$ of edge-labeling functions  $f_t: \Pi(\mathcal{Z})^{t} \rightarrow \mathcal{Y}$  such that for every path $\mu = (\mu_1, ..., \mu_d) \in \Pi(\mathcal{Z})^d$, there exists a hypothesis $h_{\mu} \in \mathcal{H}$ such that for all $t \in [d]$,  $\mathbb{E}_{z \sim \mu_t}\left[\ell(f_t(\mu_{\leq t}), z) \right] \geq \ell(f_t(\mu_{\leq t}), h_{\mu}(\mathcal{T}_t(\mu_{<t}))) + \gamma.$ The sequential minimax dimension of $\mathcal{H}$ at scale $\gamma$, denoted $\emph{\texttt{SM}}_{\gamma}(\mathcal{H}, \ell)$, is the maximal depth of a tree $\mathcal{T}$ that is $\gamma$-shattered by $\mathcal{H}$. If there exists $\gamma$-shattered trees of arbitrarily large depth, we say $\emph{\texttt{SM}}_{\gamma}(\mathcal{H}, \ell) = \infty$. Analogously, we can define $\emph{ \texttt{SM}}_{0}(\mathcal{H}, \ell)$ by requiring strict inequality, $ \mathbb{E}_{z \sim \mu}\left[\ell(f_t(\mu_{\leq t}), z) \right] > \ell(f_t(\mu_{\leq t}), h_{\mu}(\mathcal{T}_t(\mu_{<t})))$. 
\end{definition}

Observe that the SMdim is a function of both the hypothesis class $\mathcal{H}$ and the loss function $\ell$. However, when it is clear from context, we drop the dependence of $\ell$ and only write $\texttt{SM}_{\gamma}(\mathcal{H})$.  As with most scale-sensitive dimensions, SMdim has a monotonicity property, namely, $\texttt{SM}_{\gamma_1}(\mathcal{H}) \leq \texttt{SM}_{\gamma_2}(\mathcal{H}) $ for any $\gamma_2 \leq \gamma_1$. Remarkably, Theorem \ref{thm:SMdimunify} shows that the SMdim unifies all major existing results in online supervised learning. 
 
 \begin{theorem}[Unifying Learnability] \label{thm:SMdimunify}
 \noindent The following statements are true.

 \begin{itemize}
 \item[(i)] If $\mathcal{Y} = \mathcal{Z}$  and $\ell(y, z) = \mathbbm{1}\{y \neq z  \}$, then $\emph{\texttt{SM}}_{\gamma}(\mathcal{H}) = \emph{\texttt{L}}(\mathcal{H})$  for all $\gamma \in [0, \frac{1}{2}]$. 

 \item[(ii)] If $\mathcal{Y} = \mathcal{Z} = [-1, 1]$ and $\ell(y, z) = |y - z|$, then 
 $\emph{\texttt{sfat}}_{\gamma}(\mathcal{H}) \leq \emph{\texttt{SM}}_{\gamma}(\mathcal{H}) \leq \emph{\texttt{sfat}}_{\gamma'}(\mathcal{H})$ for every $0 < \gamma^{\prime} < \gamma <1$.

 \item[(ii)] If $\mathcal{Z} = \{S: S\subset \mathcal{Y}, |S| \leq k\}$ and $\ell(y, z) = \mathbbm{1}\{y \notin z\}$, then $\emph{\texttt{SM}}_{\gamma}(\mathcal{H}) = \emph{\texttt{L}}_{k+1}(\mathcal{H})$ for every $\gamma \in [0, \frac{1}{k+1}]$.

 \item[(iv)] If $\mathcal{Y} \subseteq \sigma(\Zcal)$ and $\ell(y,z) = \indicator\{z \notin y\}$, then $\emph{\texttt{SM}}_{\gamma}(\mathcal{H}) = \emph{\texttt{MS}}_{\gamma}(\mathcal{H}, \mathcal{Y})$ for all $y \in [0,1]$.
 \end{itemize}
 \end{theorem}

 Our proof of Theorem \ref{thm:SMdimunify}, found in Appendix \ref{app:SMdimunify}, uses combinatorial arguments. In particular, the proof of (ii) uses the celebrated Helly's theorem \citep{radon1921mengen}, a result from combinatorial geometry. As an immediate consequence, Theorem \ref{app:SMdimunify} shows that the SMdim provides a tight quantitative characterization of online learnability for these problems.

\section{Bounding Minimax Expected Regret} \label{sec:agn}
Our main result shows that the finiteness of SMdim at every scale $\gamma > 0$ is both necessary and sufficient for online learnability.

\begin{theorem}[Minimax Expected Regret]\label{thm:agn}
\noindent For any $(\Xcal, \Ycal, \Zcal, \Hcal, \ell)$ such that $\sup_{\gamma > 0 } \emph{\texttt{SM}}_{\gamma}(\mathcal{H}) > 0$, 
$$\sup_{\gamma > 0 } \, \gamma\,  \emph{\texttt{SM}}_{\gamma}(\mathcal{H}) \leq \inf_{\mathcal{A}} \emph{\texttt{R}}_{\mathcal{A}}(T, \mathcal{H}, \ell) \leq \inf_{\gamma  > 0 }\, \left\{\gamma T + c\, \emph{\texttt{SM}}_{\gamma}(\Hcal)+   4c\sqrt{\,\emph{\texttt{SM}}_{\gamma} (\mathcal{H})\,T \ln(T)}\right\}$$
\noindent Moreover, the upperbound and lowerbound can be tight up to logarithmic factors in $T$.
\end{theorem}

\noindent The condition $\sup_{\gamma > 0 } \texttt{SM}_{\gamma}(\mathcal{H}) > 0$ is necessary to ensure a non-negative lowerbound. \cite{raman2023online}  provide an example of a tuple $(\Xcal, \Ycal, \Zcal, \Hcal, \ell)$ with $\sup_{\gamma > 0 } \texttt{MS}_{\gamma}(\mathcal{H}) =0 $ where the corresponding minimax expected regret is negative. Moreover, \cite[Example 5.1]{raman2023online} provide a tuple $(\Xcal, \Ycal, \Zcal, \Hcal, \ell)$, where there exists an algorithm $\Acal$ such that $\inf_{\mathcal{A}} \texttt{R}_{\mathcal{A}}(T, \mathcal{H}, \ell) \leq \sup_{\gamma >0} \, \gamma\,  \texttt{MS}_{\gamma}(\Hcal)$. Since $\texttt{SM}_{\gamma}(\Hcal) = \texttt{MS}_{\gamma}(\Hcal)$ by Theorem \ref{thm:SMdimunify}, the lowerbound in Theorem \ref{thm:agn} cannot be improved in full generality. For the tightness of the upperbound, consider scalar-valued regression where $\mathcal{Y} = \Zcal = [-1,1]$, $\ell(y, z) = |y-z|$. Since, by Theorem \ref{thm:SMdimunify}, we have that $\texttt{SM}_{\gamma}(\mathcal{H}) \leq \texttt{sfat}_{\gamma^{\prime}}(\mathcal{H}) $ for all $\gamma^{\prime} < \gamma$,  Theorem \ref{thm:agn} implies that  $ \inf_{\mathcal{A}} \texttt{R}_{\mathcal{A}}(T, \mathcal{H}, \ell) \leq \inf_{\gamma>0}\{2\gamma T + 2 \texttt{sfat}_{\gamma}(\Hcal)+ 4 \sqrt{\, \texttt{sfat}_{\gamma}(\Hcal)\, T\ln(T)}\}$. However, for scalar-valued regression, \cite{rakhlin2015online} show that $\inf_{\mathcal{A}} \texttt{R}_{\mathcal{A}}(T, \mathcal{H}, \ell) \geq \sup_{\gamma>0}\,\frac{\gamma}{8}\sqrt{\texttt{sfat}_{\gamma}(\Hcal)T}$. Thus, the upperbound in Theorem \ref{thm:agn} is tight up to $O(\sqrt{\ln(T)})$.

Our proof of the lowerbound in Theorem \ref{thm:agn} is constructive. Given an algorithm and a scale $\gamma >0$, we construct a stream by traversing the sequential minimax tree of depth $\texttt{SM}_{\gamma}(\Hcal)$, adapting to the deterministic sequence of measures the algorithm uses to make its randomized prediction. Then, our claimed lowerbound follows immediately from the definition of a shattered sequential minimax tree. Since the proof of the lowerbound is relatively straightforward, we defer it to Appendix \ref{proof:agn_lower}. As for the upperbound, our proof is done in two steps. First, we consider a slight variant of the realizable setting, termed the $\varepsilon_t$-realizable setting, and show that the finiteness of SMdim at every scale $\gamma>0$ is sufficient to construct an $\varepsilon_t$-realizable learner. Finally, we convert this $\varepsilon_t$-realizable learner into an agnostic learner.

In the $\varepsilon_t$-realizable setting, an adversary plays a sequential game with the learner over $T$ rounds. In each round $t \in [T]$, the adversary selects a \textit{thresholded} labeled instance $(x_t, (y_t, \varepsilon_t)) \in \mathcal{X} \times (\mathcal{Y} \times [0, c])$ and reveals $x_t$ to the learner. The learner selects a measure $\mu_t \in \Pi(\mathcal{Z})$ and makes a randomized prediction $z_t \sim \mu_t$. Finally, the adversary reveals both the true label $y_t$ and the threshold $\varepsilon_t$ and the learner suffers the loss $\ell(y_t, z_t)$. A sequence of thresholded labeled examples $\{(x_t, (y_t, \varepsilon_t))\}_{t=1}^T$ is called $\varepsilon_t$-realizable if there exists a hypothesis $h^{\star} \in \mathcal{H}$ such that $\ell(y_t, h^{\star}(x_t)) \leq \varepsilon_t$ for all $t \in [T]$.  Given any $\varepsilon_t$-realizable stream, the goal of the learner is to output predictions such that $\sum_{t=1}^T \mathbbm{1}\{\mathbb{E}_{z \sim \mu_t}\left[\ell(y_t, z)\right] \geq \gamma + \varepsilon_t\}$ is sublinear in $T$.  We can think of the thresholds $\varepsilon_t$ as the adversary additionally revealing the loss that the best fixed hypothesis in hindsight suffers on the labeled instance $(x_t, y_t)$. This intuition is critical to our construction of an agnostic learner in Section \ref{sec:proofagn}. Note that if it is guaranteed ahead of time that $\varepsilon_t = 0$ for all $t \in [T]$, then this setting boils down to the standard realizable setting. Lemma \ref{lem:real} shows that the finiteness of $\texttt{SM}_{\gamma}(\mathcal{H})$ at every scale $\gamma > 0$ is sufficient for $\mathcal{H}$ to be online learnable in $\varepsilon_t$-realizable setting.

\begin{lemma}[$\varepsilon_t$-Realizable Learner]\label{lem:real} For any tuple $(\Xcal, \Ycal, \Zcal, \Hcal, \ell)$,  
 Algorithm \ref{alg:rand_SOA} running on any $\varepsilon_t$-realizable stream $\{(x_t, (y_t, \varepsilon_t))\}_{t=1}^T$ outputs  $\{\mu_t\}_{t=1}^T$ such that
 \begin{equation}\label{eq:randSOA}
     \sum_{t=1}^T \mathbbm{1}\Big\{\mathbb{E}_{z \sim \mu_t}\left[\ell(y_t, z)\right] \geq \gamma + \varepsilon_t \Big\} \leq  \emph{\texttt{SM}}_{\gamma}(\mathcal{H}) .
 \end{equation}
\end{lemma}

\begin{algorithm}
\setcounter{AlgoLine}{0}
\caption{Minimax Randomized Standard Optimal Algorithm (MRSOA)}\label{alg:rand_SOA}
\KwIn{$\mathcal{H}$, Target accuracy $\gamma > 0$}
Initialize $V_{0} = \mathcal{H}$

\For{$t = 1,...,T$} {
     Receive unlabeled example $x_t \in \mathcal{X}$.

     For all $(y, \varepsilon) \in \mathcal{Y} \times [0, c]$, define $V_{t-1}(y, \varepsilon) := \{h \in V_{t-1} \, \mid \, \ell(y, h(x_t)) \leq \varepsilon\}$. 

     Define $\mathcal{C}_t:= \{(y, \varepsilon) \in \mathcal{Y} \times [0, c]: |V_{t-1}(y ,\varepsilon)| > 0\}.$

    If $\texttt{SM}_{\gamma}(V_{t-1}) = 0$, pick $\mu_t \in \Pi(\Zcal)$ such that $\mathbb{E}_{z \sim \mu_t}\left[\ell(y, z)\right] < \varepsilon + \gamma$ for all $ (y, \varepsilon) \in \mathcal{C}_t$.
    
    Else, compute
    \[\mu_t = \argmin_{\mu \in \Pi(\Zcal)} \,\max_{\substack{(y, \varepsilon) \in \mathcal{Y} \times [0, c] \\ \mathbb{E}_{z \sim \mu}\left[\ell(y, z)\right] \geq \varepsilon + \gamma}} \, \texttt{SM}_{\gamma}(V_{t-1}(y, \varepsilon)).\]
    
     Predict $z_t \sim \mu_t.$
     
     Receive feedback $(y_t, \varepsilon_t)$ and update $V_t = V_{t-1}(y_t, \varepsilon_t).$

}
\end{algorithm}

To prove Lemma \ref{lem:real}, we show that (i) on any round where $\mathbb{E}_{z_t \sim \mu_t}\left[\ell(y_t, z_t) \right] \geq \gamma + \varepsilon_t$ and $\texttt{SM}_{\gamma}(V_{t-1}) > 0$, we have $\texttt{SM}_{\gamma}(V_t) \leq \texttt{SM}_{\gamma}(V_{t-1}) - 1$, and (ii) if $\texttt{SM}_{\gamma}(V_{t-1}) = 0$ there always exists a distribution $\mu_t \in \Pi(\mathcal{Z})$ such that $\mathbbm{E}_{z_t \sim \mu_t}\left[\ell(y_t, z_t)\right] < \gamma  + \varepsilon_t.$ We defer the proof to Appendix \ref{proof:lemma_RSOA}.

\subsection{Proof of Upperbound in Theorem \ref{thm:agn}}\label{sec:proofagn}
Now, we show how to convert Algorithm \ref{alg:rand_SOA} into an agnostic learner satisfying the guarantee in Theorem \ref{thm:agn}. A primary approach to proving online agnostic upperbounds involves defining a set of experts that exactly covers the hypothesis class and then running multiplicative weights \citep{cesa2006prediction} using these experts. This technique originated in \citep{ben2009agnostic} for binary classification and was later generalized by \cite{DanielyERMprinciple} to multiclass classification.  \cite{DanielyERMprinciple}'s generalization involves simulating all possible labels in $\Ycal$ to update the experts, thus making their upperbound  vacuous when $|\Ycal|$ is unbounded. Recently, \cite{hanneke2023multiclass} removed $|\Ycal|$ from the upperbound by (1) constructing an approximate cover of the hypothesis class instead of an exact cover and (2) using the feedback in the stream to update experts rather than simulating all possible labels. Our proof of the upperbound in Theorem \ref{thm:agn} combines the ideas of both \cite{DanielyERMprinciple} and \cite{hanneke2023multiclass}. In particular, following \cite{hanneke2023multiclass}, we construct an approximate cover of the hypothesis class but follow \cite{DanielyERMprinciple} in simulating all possible \textit{loss values}.\\

\begin{proof}(of upperbound in Theorem \ref{thm:agn})
    Let $(x_1, y_1), \ldots, (x_T, y_T) $ be the data stream and $h^{\star} \in \argmin_{h \in \mathcal{H}} \sum_{t=1}^T \ell(y_t, h(x_t))$ be an optimal function in hind-sight.  For a target accuracy $\gamma > 0$,  let $d_{\gamma} = \texttt{SM}_{\gamma}(\Hcal)$. Given time horizon $T$, let  $L_T = \{L \subset [T]; |L| \leq d_{\gamma}\}$ denote the set of all possible subsets of $[T]$ with size at most $d_{\gamma}$. For $\alpha \in [0,1]$, let $\{0, \alpha, \ldots, \ceil{\frac{c}{\alpha}}\alpha\}$ be an $\alpha$-cover of the loss space $[0,c]$. For every $L \in L_T$, define $\Phi_{L} = \{0, \alpha, \ldots, \ceil{\frac{c}{\alpha}}\alpha\}^L$ to be the set of all functions from $L $ to the $\alpha$-cover of $[0,c]$. Given $L \in L_T$ and $\phi_L \in \Phi_L$, define an expert $E_{L}^{\phi_L}$ such that 
\[E_L^{\phi_L}(x_t) := \text{MRSOA}_{\gamma}\Big(x_t \mid \{i, \phi_L(i)\}_{i \in L \cap [t-1]} \Big),\]
where  $ \text{MRSOA}_{\gamma}\Big(x_t \mid \{i, \phi_L(i)\}_{i \in L \cap [t-1]} \Big)$ is the prediction of the Minimax Randomized Standard Optimal Algorithm (MRSOA) running at scale $\gamma$ that has updated on thresholded labeled examples $\{(x_i, (y_i, \phi_L(i))\}_{i \in L \cap [t-1]}$.

Let $\mathcal{E} = \bigcup_{L \in L_T} \bigcup_{\phi_L \in \Phi_L} \{E^{\phi_L}_{L}\}$ denote the set of all Experts. Note that $|\mathcal{E}| = \sum_{i=0}^{d_{\gamma}} \left(\frac{2c}{\alpha} \right)^{i}\binom{T}{i} \leq \left(\frac{2cT}{\alpha}\right)^{d_{\gamma}}$. Finally, given our set of experts $\mathcal{E}$, we run the Multiplicative Weights Algorithm (MWA), denoted hereinafter as $\mathcal{A}$, over the stream $(x_1, y_1), ..., (x_T, y_T)$ with a learning rate $\eta = \sqrt{2 \ln(|\mathcal{E}|)/T}$. Let $B$ denote the random variable denoting the randomized prediction of all experts (or their corresponding randomized algorithms) . Then, conditioned on $B$,  Theorem 21.11 of \cite{ShwartzDavid} tells us that
\begin{align*}
   \sum_{t=1}^T \mathbb{E}\left[ \ell(y_t, \Acal(x_t)) \mid B\right] &\leq \inf_{E \in \mathcal{E}} \sum_{t=1}^T \ell(y_t, E(x_t)) + c\,\sqrt{2T\ln(|\mathcal{E}|)}
\end{align*}
Using $|\mathcal{E}| \leq \left(\frac{2cT}{\alpha}\right)^{d_{\gamma}}$, and taking expectations on both sides yields
\begin{equation}\label{eq:agn_intermediate}
    \mathbb{E}\left[\sum_{t=1}^T  \ell(y_t, \Acal(x_t)) \right] \leq  \expect \left[\inf_{E \in \mathcal{E}} \sum_{t=1}^T \ell(y_t, E(x_t)) \right] + c\,\sqrt{2d_{\gamma}T\ln\left(\frac{2cT}{\alpha} \right)}.
\end{equation}

Next, we show that the expected loss of the optimal expert is at most the loss of $h^{\star}$ plus a sublinear quantity. Define $\varepsilon_t := \ell(y_t, h^{\star}(x_t))$ to be the loss of the optimal hypothesis in hindsight on each round $t$. We use $\varepsilon_t$ to define a notion of mistake and use the mistake bound guarantee provided by Lemma \ref{lem:real}. 

 Define $\mu_t = \mu$$\text{-MRSOA}_{\gamma}\big(x_t \mid \{i, \phi_L(i)\}_{i \in L \cap [t-1]}\big)$ to be the measure returned by $\text{MRSOA}_{\gamma}$ (Algorithm 
\ref{alg:rand_SOA}) to make its randomized prediction given that the algorithm has updated on thresholded labeled examples $\{(x_i, (y_i, \phi_L(i)))\}_{i\in L \cap [t-1]}$. We say that $\mu\text{-MRSOA}_{\gamma}$ makes a mistake on round $t$ if $\expect_{z_t \sim \mu_t}\left[\ell(y_t, z_t) \right] \geq \ceil*{\frac{\varepsilon_t}{\alpha}}\alpha + \gamma $. As $ \ceil*{\frac{\varepsilon_t}{\alpha}}\alpha \geq \varepsilon_t$, the stream 

$$ (x_1,( y_1, \ceil*{\frac{\varepsilon_t}{\alpha}}\alpha)), \ldots, (x_T,( y_T, \ceil*{\frac{\varepsilon_t}{\alpha}}\alpha))$$

is  $\ceil*{\frac{\varepsilon_t}{\alpha}}\alpha$-realizable. Thus, with this notion of the mistake, Equation \ref{eq:randSOA} tells us that $\text{MRSOA}_{\gamma}$ makes at most $d_{\gamma}$ mistakes  on the stream $ (x_1,( y_1, \ceil*{\frac{\varepsilon_t}{\alpha}}\alpha)), \ldots, (x_T,( y_T, \ceil*{\frac{\varepsilon_t}{\alpha}}\alpha))$.

Since $\mu\text{-MRSOA}_{\gamma}$ is a deterministic mapping from the past examples to a probability measure in $\Pi(\Zcal)$, we can recursively define a sequence of time points where $\mu\text{-MRSOA}_{\gamma}$, had it run exactly on this sequence of time points,  would make mistakes at each time point. To that end, let
$$t_1 = \min \Big\{\, t \in [T]: \expect_{z_t \sim \mu_t}\left[\ell(y_t, z_t) \right] \geq \ceil*{\frac{\varepsilon_t}{\alpha}}\alpha + \gamma   \text{ where }  \mu_t=\mu\text{-MRSOA}_{\gamma}\big(x_t|\,\{\}\big)  \Big\}$$
be the earliest time point, where a fresh, unupdated copy of $\mu\text{-MRSOA}_{\gamma}$ makes a mistake if it exists. Given $t_1$, we recursively define $t_i$ for $i > 1$ as 

$$t_i = \min \left\{\,t > t_{i-1}: \expect_{z_t \sim \mu_t}\left[\ell(y_t, z_t) \right] \geq \ceil*{\frac{\varepsilon_t}{\alpha}}\alpha + \gamma  \text{ where }  \mu_t =\mu\text{-MRSOA}_{\gamma}\left(x_t \Big|\, \left\{t_j, \ceil*{\frac{\varepsilon_{t_j}}{\alpha}}\alpha\right\}_{j=1}^{i-1}\right)    \right\}$$
if it exists. That is, $t_i$ is the earliest timepoint in $[T]$ after $t_{i-1}$  where $\mu\text{-MRSOA}_{\gamma}$ having updated only on the sequence $\{(x_{t_j}, (y_{t_j}, \ceil*{\frac{\varepsilon_{t_j}}{\alpha}}\alpha))\}_{j=1}^{i-1}$ makes a mistake. We stop this process when we reach an iteration where no such time point in $[T]$ can be found where $\mu\text{-MRSOA}_{\gamma} $ makes a mistake. 

Using the definitions above, let $t_1, t_2, ..., $ denote the sequence of timepoints in $[T]$ selected via this recursive procedure.  Define $L^{\star} = \{t_1, t_2 ..., \}$ and $\phi_{L^{\star}}$ be the function such that $\phi_{L^{\star}}(t) = \ceil*{\frac{\varepsilon_t}{\alpha}}\alpha$ for each $t \in L^{\star}$.  Let $E_{L^{\star}}^{\phi_{L^{\star}}}$ be the expert parametrized by the pair $(L^{\star}, \phi_{L^{\star}})$. The expert  $E_{L^{\star}}^{\phi_{L^{\star}}}$ exists because Equation \eqref{eq:randSOA}
implies that $|L^{\star}| \leq d_{\gamma}$. By definition of the expert, we have $E_{L^{\star}}^{\phi_{L^{\star}}}(x_t) = \text{MRSOA}_{\gamma}\Big(x_t \mid \{i, \phi_{L^{\star}}(i)\}_{i \in L^{\star} \cap [t-1]} \Big)$ for all $t \in [T]$. Let us define $\mu_t^{\star} = \mu\text{-MRSOA}_{\gamma}\Big(x_t \mid \{i, \phi_{L^{\star}}(i)\}_{i \in L^{\star} \cap [t-1]} \Big)$. Using the guarantee of MRSOA (Algorithm \ref{alg:rand_SOA}), we obtain 
\begin{equation*}
    \begin{split}
      \expect \left[ \sum_{t=1}^T \ell\big(y_t, E_{L^{\star}}^{\phi_{L^{\star}}}(x_t)\big) \right]
      &= \sum_{t=1}^T \expect_{z_t \sim \mu_t^{\star}}\left[\ell\big(y_t, z_t\big) \right]\\
&\leq  \sum_{t=1}^Tc\,\mathbbm{1}\left\{\mathbb{E}_{z_t \sim \mu_t^{\star}}\left[\ell\big(y_t, z_t\big)\right] \geq  \ceil*{\frac{\varepsilon_t}{\alpha} }\alpha + \gamma \right\} +   \sum_{t=1}^T \, \left( \ceil*{\frac{\varepsilon_t}{\alpha}}\alpha + \gamma \right) \\
    &\leq c \, d_{\gamma}+  \sum_{t=1}^T \varepsilon_t + \alpha T+ \gamma T,
    \end{split}
\end{equation*}
where the final inequality uses the fact that the indicator is $1$ only on $L^{\star}$ whose size is $\leq d_{\gamma}$ and $\ceil*{\frac{\varepsilon_t}{\alpha}}\alpha \leq \varepsilon_t + \alpha$. Plugging this bound in Equation \eqref{eq:agn_intermediate}, we obtain
\begin{equation*}
    \begin{split}
       \mathbb{E}\left[\sum_{t=1}^T  \ell(y_t, \Acal(x_t)) \right] &\leq  \sum_{t=1}^T \varepsilon_t + c\,d_{\gamma }+ \alpha T + \gamma T+ 
      c\, \sqrt{2d_{\gamma}T\ln\left(\frac{2cT}{\alpha} \right)} \\
       &= \inf_{h \in \Hcal}\sum_{t=1}^T \ell(y_t, h(x_t)) + c\,d_{\gamma} + \gamma T+ 2c  + 2c\,\sqrt{ d_{\gamma} T \ln{(T)}},
    \end{split}
\end{equation*}
where we pick $\alpha = \frac{2c}{T}$ and use the fact that $ \varepsilon_t := \ell(y_t, h^{\star}(x_t))$. Finally, note that  $c\,d_{\gamma}+ 2c  + 2c\,\sqrt{ d_{\gamma} T \ln{(T)}} \leq c\, d_{\gamma} + 4c \sqrt{d_{\gamma} T \ln(T)}$. Since $\gamma > 0 $ is arbitrary, this completes our proof.
\end{proof}

\section{SMdim and the Finite Character Property} \label{sec:finitechar}

In addition to characterizing learnability, existing combinatorial dimensions in learning theory satisfy the ``Finite Character Property" (FCP) \citep{ben2019learnability, attias2023optimal}. 

\begin{definition}[Finite Character Property \citep{ben2019learnability}]\label{def:fcp}
\noindent A combinatorial dimension $\emph{\texttt{D}}(\Hcal, \ell, \gamma)$ is said to satisfy the finite character property if for every $d\in \naturals$ and $\gamma >0$, the statement $\emph{\texttt{D}}(\Hcal, \ell, \gamma) \geq d$ can be demonstrated by a finite set of domain point $X \subset \mathcal{X}$, and a finite subset of hypotheses $H \subset \mathcal{H}$.  
\end{definition}

In fact, according to \cite{ben2019learnability}, a dimension is any function $\texttt{D}$ that maps $(\Hcal, \ell)$ to $\naturals \cup \{0,\infty\}$ and satisfies the following two properties:  (1) $\mathcal{H}$ is learnable with respect to $\ell$ if and only if $\texttt{D}(\mathcal{H}, \ell) < \infty$ and (2) $\texttt{D}$ satisfies the FCP. Note that this definition of dimension differs from ours since (1) it requires $\texttt{D}$ to satisfy FCP and (2) it does not require $\texttt{D}$ to provide a quantitative characterization. 

Despite characterizing online learnability, the SMdim may not satisfy the FCP since it is defined using  trees with \textit{infinite} width. Naturally, this motivates the following question. 

\begin{center}
Under what conditions on $(\Xcal, \Ycal, \Zcal, \Hcal, \ell)$ does the SMdim satisfy the FCP? 
\end{center}

One way that the SMdim can satisfy the FCP is if it can be equivalently represented using trees with \textit{finite} width. 
For example, in Section \ref{sec:unify} we showed that the SMdim reduces to the Ldim, seq-fat dimension, $(k+1)$-Ldim, and the $p$-Set Ldim \citep[Definition 3.2]{raman2023online}, all of which are defined using finite-width trees, and thus satisfy the FCP. A unifying property in all these settings is the fact that $(\mathcal{Y}, \mathcal{Z}, \ell)$ is a \textit{Helly space}, a generalization of ``finite dimension" to abstract spaces. More formally, given any $(\mathcal{Y}, \mathcal{Z}, \ell)$, let  $\texttt{B}_{\ell}(y, r) := \{z \in \mathcal{Z}: \ell(y, z) \leq r\}$ denote the ``ball" of radius $r$ centered at $y$ induced by the loss $\ell$. Let $\texttt{B}_{\ell}(\mathcal{Y}, \mathcal{Z}) := \{\texttt{B}_{\ell}(y, r): y \in \mathcal{Y}, r \in [0, c] \}$ to be the set of all such balls. We say $(\mathcal{Y}, \mathcal{Z}, \ell)$ is a Helly space if the  \textit{Helly number} of $\texttt{B}_{\ell}(\mathcal{Y}, \mathcal{Z})$ is finite. 

\begin{definition}[Helly Number]\label{def:helly_num}
\noindent Let $S$ be a family of sets. The Helly number of $S$, denoted $\emph{\texttt{H}}(S)$, is the smallest number $p \in \mathbbm{N}$ such that for any collection of sets $C \subseteq S$ whose intersection is empty, there is a subset $C' \subset C$ of size at most $p$ whose intersection is empty.
\end{definition}

The Helly number of a set system roughly quantifies the property that every sequence of sets with empty intersection has a small sub-sequence with empty intersection. In this sense,  we use the Helly number of $\texttt{B}_{\ell}(\mathcal{Y}, \mathcal{Z})$ to quantify a notion of ``dimension" for the space $(\mathcal{Y}, \mathcal{Z}, \ell)$.

\begin{definition}[Helly Space]\label{def:hell_space}
\noindent Let $\mathcal{Z} = \mathcal{Y}$, $\ell: \mathcal{Y} \times \mathcal{Y} \rightarrow [0, c]$, and $\emph{\texttt{B}}_{\ell}(\mathcal{Y}, \mathcal{Z}) := \{\emph{\texttt{B}}_{\ell}(y, r): y \in \mathcal{Y}, r \in [0, c] \}$. Then, we say $(\mathcal{Y}, \mathcal{Z}, \ell)$ is a Helly space if and only if $\emph{\texttt{H}}(\emph{\texttt{B}}_{\ell}(\mathcal{Y}, \mathcal{Z})) < \infty$. Moreover, we define the Helly number of the space $(\mathcal{Y}, \mathcal{Z}, \ell)$ as $\emph{\texttt{H}}(\mathcal{Y}, \mathcal{Z}, \ell) := \emph{\texttt{H}}(\emph{\texttt{B}}_{\ell}(\mathcal{Y}, \mathcal{Z}))$. 
\end{definition}

It is not hard to verify that all existing work in supervised online learning theory has focused on Helly spaces.  For example, in classification with the 0-1 loss, one can verify that $\texttt{H}(\mathcal{Y}, \mathcal{Z}, \ell) = 2$. For scalar-valued regression with absolute-value loss, the celebrated Helly's theorem \citep{radon1921mengen} gives that $\texttt{H}(\mathcal{Y}, \mathcal{Z}, \ell) = 2$. More recently, \cite{raman2023online} showed that for online ranking with the 0-1 ranking loss, we have that $\texttt{H}(\mathcal{Y}, \mathcal{Z}, \ell) = 2$. Online learning settings where $\texttt{H}(\mathcal{Y}, \mathcal{Z}, \ell) \geq 3$ have also been studied. For example, in list online classification $\texttt{H}(\mathcal{Y}, \mathcal{Z}, \ell) = k + 1$ \citep{moran2023list}. In online learning with set-valued feedback \citep{raman2023online}, $\texttt{H}(\mathcal{Y}, \mathcal{Z}, \ell) = \texttt{H}(\mathcal{Y})$, where $\mathcal{Y}$ denotes an arbitrary set system defined over $\mathcal{Z}$. 

Remarkably, in all of these aforementioned settings, the combinatorial dimensions that characterize learnability are defined using trees whose width is \textit{exactly} $\texttt{H}(\mathcal{Y}, \mathcal{Z}, \ell)$. More importantly, our proofs establishing the equivalence between the SMdim and existing combinatorial dimensions crucially utilized the Helly property of $(\mathcal{Y}, \mathcal{Z}, \ell)$ to compress the infinite width trees in the definition of SMdim to finite-width trees. These facts naturally lead to the question of whether the finiteness of  $\texttt{H}(\mathcal{Y}, \mathcal{Z}, \ell)$ provides a sufficient condition under which the SMdim can be represented using finite-width trees, and more specifically, $\texttt{H}(\mathcal{Y}, \mathcal{Z}, \ell)$-width trees.

As an initial step towards answering this question, consider the $p$-shattering dimension defined in Definition \ref{def:pdim}. The central combinatorial object in this dimension is an $\mathcal{X}$-valued, $[p]$-ary tree $\mathcal{T}$, where $p \in \mathbb{N}$. In such a tree, each internal node of $\mathcal{T}$ has $p$ outgoing edges, where each edge is labeled by a tuple in $\mathcal{Y} \times [0, c]$. The tuple $(y, r)$ induces a ball $\texttt{B}_{\ell}(y, r) := \{z \in \mathcal{Z}: \ell(y, z) \leq r\}$ in the space $(\Ycal, \Zcal, \ell)$ and we further require that the collection-wise intersection of the balls induced by the tuples labeling the $p$ edges must be empty. Such a $[p]$-ary tree is shattered by a hypothesis class if for every root-to-leaf path there exists a hypothesis whose outputs on the sequence of instances lie in the balls induced by the tuples labeling the edges along the path. 

 \begin{definition}[$p$-shattering dimension]\label{def:pdim}
 \noindent Let $\ell: \mathcal{Z} \times \mathcal{Y} \rightarrow [0, c]$ be a loss function, $p \in \mathbb{N}$, and $\gamma > 0$. Let $\mathcal{T}$ be a complete $\mathcal{X}$-valued, $\left[p\right]$-ary tree of depth $d$. The tree $\mathcal{T}$ is $\gamma$-shattered by $\mathcal{H} \subseteq \Zcal^{\Xcal}$  if there exists a sequence $(f_1, ..., f_d)$ of edge-labeling functions  $f_t: \left[p\right]^{t} \rightarrow \mathcal{Y} \times [0, c]$  such that for every path $q = (q_1, ..., q_d) \in \left[p\right]^d$, we have $\bigcap_{i \in [p]} \emph{\texttt{B}}\left(f^1_t((q_{< t}, i)), f^2_t((q_{< t}, i)) + \gamma\right) = \emptyset$ and there exists a hypothesis $h_{q} \in \mathcal{H}$ such that for all $t \in [d]$,  $h_{q}(\mathcal{T}_t(q_{<t})) \in \emph{\texttt{B}}\left(f^1_t(q_{\leq t}), f^2_t(q_{\leq t})\right)$. The $p$-shattering dimension of $\mathcal{H}$ at scale $\gamma$, denoted $p\emph{\texttt{-dim}}_{\gamma}(\mathcal{H}, \ell)$, is the maximal depth of a tree $\mathcal{T}$ that is $\gamma$-shattered by $\mathcal{H}$. If there exists $\gamma$-shattered trees of arbitrarily large depth, we say $p\emph{\texttt{-dim}}_{\gamma}(\mathcal{H}, \ell) = \infty$. 
\end{definition}

Note that the tree in Definition \ref{def:pdim} is parameterized by both $p$ and $\gamma$. The number $p$ controls the width of the tree, while the number $\gamma$ is used to constrain the tuples labeling the edges. Thus, the $p\texttt{-dim}_{\gamma}(\mathcal{H})$ is defined for every $p \in \mathbb{N}$ and $\gamma \in [0, c]$. When $p = \texttt{H}(\mathcal{Y}, \mathcal{Z}, \ell)$, the $p$-dim also reduces to all existing combinatorial dimensions in their respective setting, and thus also provides a unification of supervised online learning theory. However, unlike the SMdim, the $\texttt{H}(\mathcal{Y}, \mathcal{Z}, \ell)$-dim is defined in terms of finite-width trees whenever $\texttt{H}(\mathcal{Y}, \mathcal{Z}, \ell) < \infty$.

When can the SMdim be equivalently represented using the finite-width trees in Definition \ref{def:pdim}? Lemma \ref{lem:rel_SMdim_pdim} provides a partial answer to this question by relating the SMdim and $p$-dim whenever $(\mathcal{Y}, \mathcal{Z}, \ell)$ is a Helly space. The key intuition behind the proof of Lemma \ref{lem:rel_SMdim_pdim} is that Helly spaces allows us to effectively ``compress" the infinite-width, $\Pi(\mathcal{Z})$-ary tree from the definition of SMdim to a finite-width, $\left[\texttt{H}(\mathcal{Y}, \mathcal{Z}, \ell)\right]$-ary tree according to the definition of $p$-dim. The full proof can be found in Appendix \ref{app:rel_SMdim_pdim}. 

\begin{lemma}[$\text{SMdim} \leq p\texttt{-dim}$]\label{lem:rel_SMdim_pdim}
\noindent For every $(\mathcal{X}, \mathcal{Y}, \mathcal{Z}, \mathcal{H}, \ell)$ such that $p^{\star}:= \emph{\texttt{H}}(\mathcal{Y}, \mathcal{Z}, \ell) < \infty$, we have $\emph{\texttt{SM}}_{\gamma}(\mathcal{H}) \leq p^{\star}\emph{\texttt{-dim}}_{\gamma'}(\mathcal{H})$ for all $\gamma^{\prime} < \gamma$.
\end{lemma}

Since we show that the finiteness of $\text{SMdim}$ at every scale is sufficient for online learnability, Lemma \ref{lem:rel_SMdim_pdim} implies that when $\texttt{H}(\mathcal{Y}, \mathcal{Z}, \ell) < \infty$, the finiteness of $p^{\star}\texttt{-dim}_{\gamma}(\mathcal{H})$ at every scale $\gamma$ is sufficient for online learnability. The following open question asks whether the finiteness of $p^{\star}\texttt{-dim}_{\gamma}(\mathcal{H})$ at every scale $\gamma > 0$ is also \textit{necessary} for online learnability.

\begin{center}
    \noindent Suppose that $p^{\star}:=\texttt{H}(\mathcal{Y}, \mathcal{Z}, \ell) < \infty$. Does online learnability of $\mathcal{H}$ imply that $p^{\star}\texttt{-dim}_{\gamma}(\mathcal{H}) < \infty$ for all $\gamma > 0$? 
\end{center}

\noindent One way to resolve this question would be to show that $p^{\star}\texttt{-dim}_{\gamma}(\mathcal{H})\leq \texttt{SM}_{\gamma}(\mathcal{H})$ for all $\gamma > 0$. A positive resolution ultimately implies that $(\mathcal{Y}, \mathcal{Z}, \ell)$ being a Helly space is a sufficient condition for SMdim to be equivalently represented using finite-width trees and therefore satisfy the FCP.

\bibliographystyle{plainnat}
\bibliography{references}

\newpage 
\appendix

\section{Proof of Theorem \ref{thm:SMdimunify}} \label{app:SMdimunify}

In this section, we show that SMdim reduces to existing combinatorial dimensions. We start with Lemma \ref{lem:GMS=Ldim}, which shows that SMdim $\equiv$ Ldim. 

\begin{lemma}[SMdim $\equiv$ Ldim]\label{lem:GMS=Ldim} 
    \noindent Let $\mathcal{Y} = \mathcal{Z}$,  $\mathcal{H} \subseteq \mathcal{Z}^{\mathcal{X}}$, and $\ell(y, z) = \mathbbm{1}\{y \neq z  \}$. Then, $\emph{\texttt{SM}}_{\gamma}(\mathcal{H}) = \emph{\texttt{L}}(\mathcal{H})$ for all $\gamma \in [0, \frac{1}{2}]$. 
\end{lemma}

\begin{proof}
 Fix $\gamma \in (0, \frac{1}{2}]$. We first show that $\texttt{SM}_{\gamma}(\mathcal{H}) \leq \texttt{L}(\mathcal{H})$. Let $\mathcal{T}$ be a $\mathcal{X}$-valued, $\Pi(\mathcal{Z})$-ary tree of depth $d = \texttt{SM}_{\gamma}(\mathcal{H})$ shattered by $\mathcal{H}$. Let $v$ be the root node of $\mathcal{T}$ and $x$ denote the instance labeling the node. Recall that $v$ has an outgoing edge for each measure $\mu \in \Pi(\mathcal{Z})$. Let $\{y_{\mu}\}_{\mu \in \Pi(\mathcal{Z})}$ be the set of elements in $\mathcal{Y}$ that label the outgoing edges from $v$. We first claim that there at least two distinct elements in the set $\{y_{\mu}\}_{\mu \in \Pi(\mathcal{Z})}$. For the sake of contradiction, suppose this is not the case. That is, there is only one distinct element that labels the outgoing edges from $v$. Let $y $ denote the element that labels the outgoing edges from $v$. That is, $y_{\mu} = y$ for all $\mu \in \Pi(\mathcal{Z})$. Consider the Dirac measure $\delta_y$ that puts all mass on $y$. Note that $\delta_y \in \Pi(\mathcal{Z})$ and therefore there exists an outgoing edge from $v$ indexed by $\delta_y$ and labeled by $y$. However, it must be the case that $\mathbb{P}_{z \sim \delta_y}\left[y \neq z \right] = 0$. Since $\gamma > 0$, the shattering condition required by Definition \ref{def:SMdim} cannot be met, which is a contradiction. Accordingly, there is at least two distinct elements in the set $\{y_{\mu}\}_{\mu \in \Pi(\mathcal{Z})}$. 

Let $y_{-1}, y_{+1}$ be the distinct elements of the set $\{y_{\mu}\}_{\mu \in \Pi(\mathcal{Z})}$, and $\mu_{-1}, \mu_{+1}$ be the indices of the edges labeled by $y_{-1}$ and $y_{+1}$ respectively. Let $\mathcal{H}_{-1} = \{h_\mu: \mu \in \Pi(\mathcal{Z})^d, \mu_1 = \mu_{-1}\}$ denote the set of shattering hypothesis that corresponds to following a path down $\mathcal{T}$ that takes the outgoing edge indexed $\mu_{-1}$ from the root node. Likewise define $\mathcal{H}_{+1}$. Keep the edges indexed by $\mu_{-1}$ and $\mu_{+1}$ and remove all other outgoing edges along with their corresponding subtree. Reindex the two edges using $\{\pm 1\}$. The root node $v$ should now have two outgoing edges, indexed by $\{\pm 1\}$ and labeled by distinct elements of $\mathcal{Y}$, matching the first constraint of a Littlestone tree. As for the second constraint, observe that for all $h_{-1} \in \mathcal{H}_{-1}$ and $h_{+1} \in \mathcal{H}_{+1}$  the shattering condition from Definition \ref{def:SMdim} implies that $\mathbb{P}_{z \sim \mu_{-1}}\left[y_{-1} \neq z \right] \geq \mathbbm{1}\{y_{-1} \neq h_{-1}(x)\} + \gamma$ and $\mathbb{P}_{z \sim \mu_{+1}}\left[y_{+1} \neq z \right] \geq \mathbbm{1}\{y_{+1} \neq h_{+1}(x)\} + \gamma.$  However, this can only be true if both $\mathbbm{1}\{y_{-1} \neq h_{-1}(x)\} = 0 \implies y_{-1} = h_{-1}(x)$ and $\mathbbm{1}\{y_{+1} \neq h_{+1}(x)\} = 0 \implies y_{+1} = h_{+1}(x)$. Accordingly, the hypotheses that shatters the edges indexed by $\mu_{-1}$ and $\mu_{+1}$ in the original tree according to Definition \ref{def:SMdim} also shatters the newly re-indexed edges according to Definition \ref{def:ldim}. Recursively repeating the above procedure on the subtrees following the two reindexed edges results in a Littlestone tree shattered by $\mathcal{H}$ of depth $d$. Thus, $\texttt{SM}_{\gamma}(\mathcal{H}) \leq \texttt{L}(\mathcal{H})$ for $\gamma \in (0, \frac{1}{2}]$. The case when $\gamma = 0$ follows similarly and uses the fact that when $\gamma = 0$, we define the shattering condition in SMdim with a strict inequality (see last sentence in Definition \ref{def:SMdim}).

We now prove the inequality that $\texttt{SM}_{\gamma}(\mathcal{H}) \geq \texttt{L}(\mathcal{H})$. Fix $\gamma \in [0, \frac{1}{2}]$. Let $\mathcal{T}$ be a $\mathcal{X}$-valued, $\{\pm1\}$-ary tree of depth $d = \texttt{L}(\mathcal{H})$ shattered by $\mathcal{H}$ according to Definition \ref{def:ldim}. Our goal will be to expand $\mathcal{T}$ into a $\Pi(\mathcal{Z})$-ary tree that is $\gamma$-shattered by $\mathcal{H}$ according to Definition \ref{def:SMdim}. Let $v$ be the root node of $\mathcal{T}$, $x$ be the instance that labels the root node, and $y_{-1}, y_{+1}$ denote the distinct elements of $\mathcal{Y}$ that label the left and right outgoing edges from $v$ respectively.  Let $\mathcal{H}_{-1} = \{h_\sigma: \sigma \in \{\pm1\}^d, \sigma_1 = -1\} \subset \mathcal{H}$ denote the set of shattering hypothesis that correspond to following a path down $\mathcal{T}$ that takes the edge indexed by $-1$ in the first level. Define $\mathcal{H}_{+1}$ analogously. Then, for all $h_{-1} \in \mathcal{H}_{-1}$ and $h_{+1} \in \mathcal{H}_{+1}$,  the shattering condition implies that $h_{-1}(x) = y_{-1}$ and $h_{+1}(x) = y_{+1}$.

For every measure $\mu \in \Pi(\mathcal{Z})$, we claim that there exists a $\sigma_{\mu} \in \{\pm1\}$ such that $\mathbb{P}_{z \sim \mu}\left[y_{\sigma_{\mu}} \neq z \right] = \mu(\{y_{\sigma_{\mu}}\}^c) \geq \gamma$. Suppose for the sake of contradiction that this is not true. Then, there exists a measure $\mu \in \Pi(\mathcal{Z})$ such that for both $\sigma \in \{\pm1\}$, we have $\mu(\{y_{\sigma}\}^c) < \gamma$. Then, $1 = \mu(\mathcal{Z}) = \mu(\{y_{-1}\}^c \cup \{y_{+1}\}^c) < 2\gamma < 1$, a contradiction. Thus, for every measure $\mu \in \Pi(\mathcal{Z})$ there exists a $\sigma_{\mu} \in \{\pm1\}$ such that $\mathbb{P}_{z \sim \mu}\left[y_{\sigma_{\mu}} \neq z \right] \geq \gamma$. Combining this with the fact that for any $h_{-1} \in \mathcal{H}_{-1}$ and  $h_{+1} \in \mathcal{H}_{+1}$, we have $y_{-1} = h_{-1}(x)$ and  $y_{+1} = h_{+1}(x)$,  gives that, for every measure $\mu \in \Pi(\mathcal{Z})$, there exists a $\sigma_{\mu} \in \{\pm1\}$ such that for all $h_{\sigma_{\mu}} \in \mathcal{H}_{\sigma_{\mu}}$ ,we have $\mathbb{P}_{z \sim \mu}\left[y_{\sigma_{\mu}} \neq z \right] \geq \mathbbm{1}\{y_{\sigma_{\mu}} \neq h_{\sigma_{\mu}}(x)\} + \gamma$. Note that if we take $y_{\sigma_{\mu}}$ to be the label on an edge indexed by $\mu$, then the inequality above matches the shattering condition required by Definition \ref{def:SMdim}. 

To that end, for every measure $\mu \in \Pi(\mathcal{Z})$, add an outgoing edge from $v$ indexed by $\mu$ and labeled by the $y_{\sigma_{\mu}}$, where $\sigma_{\mu}$ is the index as promised by the analysis above. Grab the sub-tree in $\mathcal{T}$ following the original outgoing edge from $v$ indexed by $\sigma_{\mu}$, and append it to the newly constructed outgoing edge from $v$ indexed by $\mu$. Remove the original outgoing edges from $v$ indexed by $\{\pm1\}$ and their corresponding subtrees. Recursively repeat the above procedure on the subtrees following the newly created edges indexed by measures. Upon repeated this process for every internal node in $\mathcal{T}$, we obtain a $\Pi(\mathcal{Z})$-ary tree that is $\gamma$-shattered by $\mathcal{H}$ of depth $d$. Thus, we have that $\texttt{L}(\mathcal{H}) \leq \texttt{SM}_{\gamma}(\mathcal{H})$ for $\gamma \in [0, \frac{1}{2}]$. 
\end{proof}

Next, we show an equivalence between SMdim and seq-fat. 

\begin{lemma}[SMdim $\equiv$ seq-fat]\label{lem:fateq}
    \noindent Let $\mathcal{Y} = \mathcal{Z} = [-1, 1]$,  $\mathcal{H} \subseteq \mathcal{Z}^{\mathcal{X}}$, and $\ell(y, z) = |y - z|$. Then for every $\gamma \in (0, 1]$ and $\gamma' < \gamma$, 

    $$\emph{\texttt{sfat}}_{\gamma}(\mathcal{H}) \leq \emph{\texttt{SM}}_{\gamma}(\mathcal{H}) \leq \emph{\texttt{sfat}}_{\gamma'}(\mathcal{H}).$$    
\end{lemma}

\begin{proof}
We first prove the upperbound. Let $\gamma \in (0, 1]$ and $\gamma' < \gamma$. Let $\mathcal{T}$ be a $\mathcal{X}$-valued, $\Pi(\mathcal{Z})$-ary tree of depth $d = \texttt{SM}_{\gamma}(\mathcal{H})$ shattered by $\mathcal{H}$. Let $v$ be the root node of $\mathcal{T}$ and $x$ denote the instance labeling the node. Recall that $v$ has an outgoing edge for each measure $\mu \in \Pi(\mathcal{Z})$. In particular, this means that $v$ has outgoing edges corresponding to the Dirac measures on $\mathcal{Z}$, which we denote by $\{\delta_z\}_{z \in \mathcal{Z}}$.  Fix a $z \in \mathcal{Z}$  and consider the  outgoing edge from $v$ indexed by $\delta_z$. Let $y_{z} \in \mathcal{Y}$ be the element that labels the outgoing edge indexed by $\delta_{z}$. Let $\mathcal{H}_{z} = \{h_{\mu}: \mu \in \Pi(\mathcal{Z})^d,  \mu_1 = \delta_{z}\} \subset \mathcal{H}$ denote the set of shattering hypothesis that corresponds to following a path down $\mathcal{T}$ that takes the edge $\delta_z$ in the root node. Then, for all $h \in \mathcal{H}_{z}$ the shattering condition from Definition \ref{def:SMdim} implies that

$$|z - y_{z}| \geq |h(x) - y_{z}| + \gamma > |h(x) - y_{z}| + \gamma'.$$

Taking the supremum on both sides, gives that:

\begin{equation}\label{eq:relation_ineq}
|z - y_{z}| > \sup_{h \in \mathcal{H}_{z}}|h(x) - y_{z}| + \gamma' = r_z + \gamma'.
\end{equation}

where we let $r_z = \sup_{h \in \mathcal{H}_{z}}|h(x) - y_{z}|$.
Let $I_z := \left[y_{z} - (r_z + \gamma'), y_{z} + (r_z + \gamma')\right] \subset [-3, 3]$ denote an interval corresponding to $z$.  Inequality \eqref{eq:relation_ineq} above implies that $z \notin I_z$ (note that $I_z$ changes depending on $z$). Since $z \in \mathcal{Z}$ was arbitrary, it must be the case that $z \notin I_z$ for all $z \in \mathcal{Z}$. This means that $\bigcap_{z \in \mathcal{Z}} I_z = \emptyset$.   Since $[-3, 3]$ is compact and $\{I_z\}_{z \in \mathcal{Z}}$ is a family of closed intervals whose intersection is empty, the celebrated Helly's theorem states that there exists two intervals in $\{I_z\}_{z \in \mathcal{Z}}$ that are disjoint \citep{ECKHOFF1993389, radon1921mengen}. Accordingly, let $z_1, z_2$ be such that $I_{z_1} \cap I_{z_2} = \emptyset$. As before, let $y_{z_1}$ and $y_{z_2}$ be the labels on the outgoing edges from $v$ indexed by the Dirac measures $\delta_{z_1}$ and $\delta_{z_2}$ respectively. Without loss of generality, let $y_{z_1} < y_{z_2}$ (we have strict inequality because we are guaranteed that $I_{z_1}$ and $I_{z_2}$ are disjoint). By inequality \ref{eq:relation_ineq}, for all $h_{z_1} \in \mathcal{H}_{z_1}$ and $h_{z_2} \in \mathcal{H}_{z_2}$ we have that

$$h_{z_1}(x) \in \left[y_{z_1} - r_{z_1}, y_{z_1} + r_{z_1} \right] \quad\text{and}\quad h_{z_2}(x) \in \left[y_{z_2} - r_{z_2}, y_{z_2} + r_{z_2} \right].$$

 Let $s = \frac{y_{z_1} + r_{z_1} + y_{z_2} - r_{z_2}}{2} \in [-1, 1]$ be a witness. Then, for all $h_{z_1} \in \mathcal{H}_{z_1}$ and $h_{z_2} \in \mathcal{H}_{z_2}$,  we have that $s - h_{z_1}(x) \geq \gamma'$ and $h_{z_2}(x) - s \geq \gamma'$.  Relabel the two edges indexed by $\delta_{z_1}$ and $\delta_{z_2}$ with the same witness $s$. Reindex the two edges indexed by $\delta_{z_1}$ and $\delta_{z_2}$ with $-1$ and $+1$ respectively. Remove all other edges indexed by measures and their corresponding subtrees. There should now only be two outgoing edges from $v$, each labeled by the same witness. Next, recall that for all $h_{z_1} \in \mathcal{H}_{z_1}$ and $h_{z_2} \in \mathcal{H}_{z_2}$ we have that $s - h_{z_1} \geq \gamma'$ and $h_{z_2} - s \geq \gamma'$. Accordingly, the hypotheses that shatter the  edges indexed by $\delta_{z_1}$ and $\delta_{z_2}$ in the original tree according to Definition \ref{def:SMdim} also shatter the newly re-indexed and relabeled edges according to Definition \ref{def:sfat}. Recursively repeating the above procedure on the subtrees following the two newly reindexed and relabeled edges results in a seq-fat tree $\gamma'$-shattered by $\mathcal{H}$ of depth $d$. Thus, 
$\texttt{SM}_{\gamma}(\mathcal{H}) \leq \texttt{sfat}_{\gamma'}(\mathcal{H})$ for $\gamma' < \gamma$. 

We now move on to prove the lowerbound.  Let $\gamma \in (0, 1]$ and $\mathcal{T}$ be a $\mathcal{X}$-valued, $\{\pm1\}$-ary tree of depth $d = \texttt{sfat}_{\gamma}(\mathcal{H})$ shattered by $\mathcal{H}$ according to Definition \ref{def:sfat}. Our goal will be expand $\mathcal{T}$ into a $\Pi(\mathcal{Z})$-ary tree that is $\gamma$-shattered by $\mathcal{H}$ according to Definition \ref{def:SMdim}. Let $v$ be the root node, $x$ the instance that labels the root node, and $s$ be the witness that labels the two outgoing edges of $v$. Let $\mathcal{H}_{-1} = \{h_{\sigma}: \sigma \in \{\pm1\}^{d}, \sigma_1 = -1\} \subset \mathcal{H}$ denote the set of shattering hypothesis that corresponds to following a path down $\mathcal{T}$ that takes the outgoing edge indexed by $-1$ from the root node. Likewise define $\mathcal{H}_{+1}$. Then, for all $h_{-1} \in \mathcal{H}_{-1}$ and $h_{+1} \in \mathcal{H}_{+1}$, the shattering condition implies that $ s - h_{-1}(x) \geq \gamma$ and $h_{+1}(x) - s \geq \gamma$ respectively.  


For every measure $\mu \in \Pi(\mathcal{Z})$,  we claim that there exist a $\sigma_{\mu} \in \{-1, 1\}$ such that $\mathbb{E}_{z \sim \mu}\left[|\sigma_{\mu} - z| \right] \geq |s - \sigma_{\mu}|$. Suppose for the sake of contradiction that this is not true. That is, there exists $\mu \in \Pi(\mathcal{Z})$ such that for all $\tau \in \{-1, 1\}$ we have that $\mathbb{E}_{z \sim \mu}\left[|\tau - z| \right] < |s - \tau|$. Then, when $\tau = -1$, we have that $\mathbb{E}_{z \sim \mu}\left[z\right] < |s + 1| - 1$ and when $\tau = 1$, we have $1 - |s - 1| < \mathbb{E}_{z \sim \mu}\left[z\right]$, using the fact that $|\tau - z| = 1 -\tau z $. Combining the two inequalities together and using the fact that $s \in [-1, 1]$ gives that $2 < |s + 1| + |s - 1| = 2$, which is a contradiction.  Accordingly, for every measure $\mu \in \Pi(\mathcal{Z})$, there exists a $\sigma_{\mu} \in \{-1, 1\}$ such that $\mathbb{E}_{z \sim \mu}\left[|\sigma_{\mu} - z| \right] \geq |s - \sigma_{\mu}|$. Next, crucially note that for any $\tau \in \{\pm1\}$ and any $h_{\tau} \in \mathcal{H}_{\tau}$, we have $|h_{\tau}(x) - \tau| = |s - \tau| - |h_{\tau}(x) - s| \leq |s - \tau| - \gamma$ by the seq-fat shattering condition from Definition \ref{def:sfat}. Therefore, for every measure $\mu \in \Pi(\mathcal{Z})$, there exists  $\sigma_{\mu} \in \{\pm1\}$ such that for all $h_{\sigma_{\mu}} \in \mathcal{H}_{\sigma_{\mu}}$, we have that $\mathbb{E}_{z \sim \mu}\left[|\sigma_{\mu} - z| \right] \geq |\sigma_{\mu} - h_{\sigma_{\mu}}(x)| + \gamma$. Note that if we take $\sigma_{\mu}$ to be the label on a edge indexed by $\mu$, then $\mathbb{E}_{z \sim \mu}\left[|\sigma_{\mu} - z| \right] \geq |\sigma_{\mu} - h_{\sigma_{\mu}}(x)| + \gamma$ exactly matches the shattering condition required by Definition \ref{def:SMdim}. 

To that end, for every measure $\mu \in \Pi(\mathcal{Z})$, add an outgoing edge from $v$ indexed by $\mu$ and labeled by the $\sigma_{\mu} \in \{\pm1\}$ promised in the analysis above. Grab the sub-tree in $\mathcal{T}$ following the original outgoing edge from $v$ indexed by $\sigma_{\mu}$, and append it to the newly constructed outgoing edge from $v$ indexed by $\mu$. Remove the original outgoing edges from $v$ indexed by $-1$ and $+1$ and their corresponding subtrees. Recursively repeat the above procedure on the subtrees following the newly created edges indexed by measures. Upon repeating this process for every internal node in $\mathcal{T}$, we obtain a $\Pi(\mathcal{Z})$-ary tree that is $\gamma$-shattered by $\mathcal{H}$ of depth $d$. Thus, we have that $\texttt{sfat}_{\gamma}(\mathcal{H}) \leq \texttt{SM}_{\gamma}(\mathcal{H})$. 
\end{proof}

Next, we show that SMdim reduces to $(k+1)$-Ldim from \cite{moran2023list}. 

\begin{lemma}[SMdim $\equiv$ $(k+1)$-Ldim]{}
    \noindent Let $\mathcal{Z} = \{S: S\subset \mathcal{Y}, |S| \leq k\}$,  $\mathcal{H} \subseteq \mathcal{Z}^{\mathcal{X}}$, and $\ell(y, z) = \mathbbm{1}\{y \notin z\}$. Then for every $\gamma \in [0, \frac{1}{k+1}]$, we have $\emph{\texttt{SM}}_{\gamma}(\mathcal{H}) = \emph{\texttt{L}}_{k+1}(\mathcal{H}).$   
\end{lemma}

\begin{proof}
 Fix $\gamma \in (0, 1]$. We first show that $\texttt{SM}_{\gamma}(\mathcal{H}) \leq \texttt{L}_{k+1}(\mathcal{H})$. Let $\mathcal{T}$ be a $\mathcal{X}$-valued, $\Pi(\mathcal{Z})$-ary tree of depth $d = \texttt{SM}_{\gamma}(\mathcal{H})$ shattered by $\mathcal{H}$. Let $v$ be the root node of $\mathcal{T}$ and $x$ denote the instance labeling the node. Recall that $v$ has an outgoing edge for each measure $\mu \in \Pi(\mathcal{Z})$. Let $\{y_{\mu}\}_{\mu \in \Pi(\mathcal{Z})}$ be the set of elements in $\mathcal{Y}$ that label the outgoing edges from $v$. We first claim that there at least $k+1$ distinct elements in the set $\{y_{\mu}\}_{\mu \in \Pi(\mathcal{Z})}$. For the sake of contradiction, suppose this was not the case. That is, there are only $k$ distinct elements that label the outgoing edges from $v$. Let $y_1, ..., y_k$ denote the $k$ distinct elements that label the outgoing edges from $v$, Consider the measure $\tilde{\mu}$ that puts all mass on $\{y_1, ..., y_k\}$. Note that $\tilde{\mu} \in \Pi(\mathcal{Z})$ and let $\tilde{y} \in \{y_1, ..., y_k\}$ be the label on the outgoing edge from $v$ indexed by $\tilde{\mu}$. By definition of $\tilde{\mu}$ and $\tilde{y}$, it must be the case that $\mathbb{P}_{z \sim \tilde{\mu}}\left[\tilde{y} \notin z \right] = 0$. Since $\gamma > 0$, the shattering condition required by Definition \ref{def:SMdim} cannot be met, which is a contradiction. Accordingly, there exists at least $k+1$ distinct elements in the set $\{y_{\mu}\}_{\mu \in \Pi(\mathcal{Z})}$. 

Let $y_1, ..., y_{k+1}$ be the distinct elements of the set $\{y_{\mu}\}_{\mu \in \Pi(\mathcal{Z})}$, and $\mu_1, ..., \mu_{k+1}$ be the indices of the edges labeled by $y_1, ..., y_{k+1}$ respectively, breaking ties arbitrarily. For $\mu_i \in \{\mu_1, ..., \mu_{k+1}$\}, let $\mathcal{H}_{\mu_i}$ denote the set of shattering hypothesis that corresponds to following a path down $\mathcal{T}$ that takes the outgoing edge $\mu_i$ from the root node. Keep the edges indexed by $\mu_1, ..., \mu_{k+1}$, and remove all other outgoing edges along with their corresponding subtree. Reindex the $k+1$ edges using distinct numbers in $[k+1]$. The root node $v$ should now have $k+1$ outgoing edges, each indexed by a different natural number in $[k+1]$ and labeled by a distinct element of $\mathcal{Y}$, matching the first constraint of a $(k+1)$-Littlestone tree. As for the second constraint, observe that for all $h \in \mathcal{H}_{\mu_i}$ the shattering condition implies that $\mathbb{P}_{z \sim \mu_i}\left[y_i \notin z \right] \geq \mathbbm{1}\{y_i \notin h(x)\} + \gamma.$ However, this can only be true if $\mathbbm{1}\{y_i \notin h(x)\} = 0 \implies y_i \in h(x)$. Accordingly, the hypotheses that shatter the edges indexed by $\mu_1, ..., \mu_{k+1}$ in the original tree according to Definition \ref{def:SMdim} also shatter the newly re-indexed edges according to Definition \ref{def:kldim}. Recursively repeating the above procedure on the subtrees following the $k+1$ reindexed edges results in a $(k+1)$-Littlestone tree shattered by $\mathcal{H}$ of depth $d$. Thus, $\texttt{SM}_{\gamma}(\mathcal{H}) \leq \texttt{L}_{k+1}(\mathcal{H})$ for $\gamma \in (0, 1]$. The case when $\gamma = 0$ follows similarly and uses the fact that when $\gamma = 0$, we define the shattering condition in SMdim with a strict inequality (see last sentence in Definition \ref{def:SMdim}).

We now prove the inequality that $\texttt{SM}_{\gamma}(\mathcal{H}) \geq \texttt{L}_{k+1}(\mathcal{H})$. Fix $\gamma \in [0, \frac{1}{k+1}]$.  Let $\mathcal{T}$ be a $\mathcal{X}$-valued, $[k+1]$-ary tree of depth $d = \texttt{L}_{k+1}(\mathcal{H})$ shattered by $\mathcal{H}$ according to Definition \ref{def:kldim}. Our goal will be to expand $\mathcal{T}$ into a $\Pi(\mathcal{Z})$-ary tree that is $\gamma$-shattered by $\mathcal{H}$ according to Definition \ref{def:SMdim}. Let $v$ be the root node of $\mathcal{T}$, $x$ be the instance that labels the root node, and $\{y_i\}_{i=1}^{k+1}$ denote the distinct elements of $\mathcal{Y}$ that label the $k+1$ outgoing edges from $v$. For each $i \in [k+1]$, let $\mathcal{H}_i = \{h_p: p \in [k+1]^{d}, p_1 = i\} \subset \mathcal{H}$ denote the set of shattering hypothesis that corresponds to following a path down $\mathcal{T}$ that takes the outgoing edge indexed by $i$ from $v$. Then, for all $i \in [k+1]$ and $h_i \in \mathcal{H}_i$,  the shattering condition implies that $y_i \in h_i(x) \implies \mathbbm{1}\{y_i \notin h_i(x)\} = 0$.

For every measure $\mu \in \Pi(\mathcal{Z})$, we claim that there exists a $i_{\mu} \in [k+1]$ such that $\mathbb{P}_{z \sim \mu}\left[y_{i_{\mu}} \notin z \right] \geq \gamma$. Suppose for the sake of contradiction that this is not true. Then, there exists a measure $\mu \in \Pi(\mathcal{Z})$ such that for all $i \in [k+1]$, we have $\mathbb{P}_{z \sim \mu}\left[y_{i} \notin z \right] < \gamma$. This implies that 
$$\mathbb{P}_{z \sim \mu}\left[\exists i \in [k+1] \text{ such that } y_{i} \notin z \right] < (k+1)\gamma < 1.$$ 
However, since $\mu$ is supported over subsets of $\mathcal{Y}$ of size  $\leq k$, we have $\mathbb{P}_{z \sim \mu}\left[\exists i \in [k+1] \text{ such that } y_{i} \notin z \right] = 1$, a contradiction. Thus, for every measure $\mu \in \Pi(\mathcal{Z})$ there exists a $i_{\mu} \in [k+1]$ such that $\mathbb{P}_{z \sim \mu}\left[y_{i_{\mu}} \notin z \right] \geq \gamma$. Combining this with the fact that for every $i \in [k+1]$ and $h_i \in \mathcal{H}_i$ we have that $y_i \in h_i(x)$ gives that, for every measure $\mu \in \Pi(\mathcal{Z})$, there exists a $i_{\mu} \in [k+1]$ such that for all $h_{i_{\mu}} \in \mathcal{H}_{i_{\mu}}$, we have $\mathbb{P}_{z \sim \mu}\left[y_{i_{\mu}} \notin z \right] \geq \mathbbm{1}\{y_{i_{\mu}} \notin h_{i_{\mu}}(x)\} + \gamma$. Note that if we take $y_{i_{\mu}}$ to be the label on an edge indexed by $\mu$, then the inequality above matches the shattering condition required by Definition \ref{def:SMdim}. 

To that end, for every measure $\mu \in \Pi(\mathcal{Z})$, add an outgoing edge from $v$ indexed by $\mu$ and labeled by the $y_{i_{\mu}}$, where $i_{\mu}$ is the index as promised by the analysis above. Grab the sub-tree in $\mathcal{T}$ following the original outgoing edge from $v$ indexed by $i_{\mu}$, and append it to the newly constructed outgoing edge from $v$ indexed by $\mu$. Remove the original outgoing edges from $v$ indexed by numbers in $[k+1]$ and their corresponding subtrees. Recursively repeat the above procedure on the subtrees following the newly created edges indexed by measures. Upon repeating this process for every internal node in $\mathcal{T}$, we obtain a $\Pi(\mathcal{Z})$-ary tree of depth $d$ that is $\gamma$-shattered by $\mathcal{H}$. Thus, we have that $\texttt{L}_{k+1}(\mathcal{H}) \leq \texttt{SM}_{\gamma}(\mathcal{H})$ for $\gamma \in [0, \frac{1}{k+1}]$. 
\end{proof}

Finally, we show that the SMdim $\equiv$ MSdim. 

\begin{lemma}[SMdim $\equiv$ MSdim]
    \noindent Let $\mathcal{Y} \subset \sigma(\mathcal{Z})$,  $\mathcal{H} \subseteq \mathcal{Z}^{\mathcal{X}}$, and $\ell(y, z) = \mathbbm{1}\{z \notin y\}$. Then for every $\gamma \in [0, 1]$, we have $\emph{\texttt{SM}}_{\gamma}(\mathcal{H}) = \emph{\texttt{MS}}_{\gamma}(\mathcal{H}).$   
\end{lemma}

\begin{proof}
    The equality follows directly from the fact that $\mathbb{E}_{z \sim \mu}\left[\ell(y, z)\right] = \mu(y^c)$ and the fact that $\mathbb{E}_{z \sim \mu_t}\left[\ell(z, f_t(\mu_{\leq t})) \right] \geq \ell(h_{\mu}(\mathcal{T}_t(\mu_{<t})),  f_t(\mu_{\leq t})) + \gamma \iff h_{\mu}(\mathcal{T}_t(\mu_{<t})) \in f_t(\mu_{\leq t}) \text{ and } \mu_t(f_t(\mu_{\leq t})) \leq 1 - \gamma$.  
\end{proof}

\section{Proof of Lemma \ref{lem:real}}\label{proof:lemma_RSOA}

We now prove that given any target accuracy $\gamma > 0$ and any $\varepsilon_t$-realizable sequence $\{(x_t, (y_t, \varepsilon_t))\}_{t=1}^T$, Algorithm \ref{alg:rand_SOA} computes distributions $\mu_t \in \Pi(\mathcal{Z})$ such that
\[
\sum_{t=1}^T \mathbbm{1}\{\mathbb{E}_{z \sim \mu_t}\left[\ell(y_t, z)\right] \geq \gamma + \varepsilon_t\} \leq \texttt{SM}_{\gamma}(\mathcal{H}).
\]

\noindent To prove this guarantee, it suffices to show that (i) on any round where $\mathbb{E}_{z_t \sim \mu_t}\left[\ell(y_t, z_t) \right] \geq \gamma + \varepsilon_t$ and $\texttt{SM}_{\gamma}(V_{t-1}) > 0$, we have $\texttt{SM}_{\gamma}(V_t) \leq \texttt{SM}_{\gamma}(V_{t-1}) - 1$, and (ii) if $\texttt{SM}_{\gamma}(V_{t-1}) = 0$ there always exists a distribution $\mu_t \in \Pi(\mathcal{Z})$ such that $\mathbbm{E}_{z_t \sim \mu_t}\left[\ell(y_t, z_t)\right] < \gamma  + \varepsilon_t.$

Let $t \in [T]$ be a round where $\mathbb{E}_{z_t \sim \mu_t}\left[\ell(y_t, z_t) \right] \geq \gamma + \varepsilon_t$ and $\texttt{SM}_{\gamma}(V_{t-1}) > 0$.  For the sake contradiction, suppose that $\texttt{SM}_{\gamma}(V_t) = \texttt{SM}_{\gamma}(V_{t-1}) = d$. Then, by the min-max computation in Algorithm \ref{alg:rand_SOA}, for every measure $\mu \in \Pi(\mathcal{Z})$, there exists a pair $(y_{\mu}, \varepsilon_{\mu}) \in \mathcal{Y} \times [0, c]$ such that $\mathbb{E}_{z \sim \mu}\left[\ell(y_{\mu}, z)\right] \geq \varepsilon_{\mu} + \gamma $ and $\texttt{SM}_{\gamma}(V_{t-1}(y_{\mu}, \varepsilon_{\mu})) = d$. Now construct a tree $\mathcal{T}$ with $x_t$ labeling the root node. For each measure $\mu \in \Pi(\mathcal{Z})$, construct an outgoing edge from $x_t$ indexed by $\mu$ and  labeled by $y_{\mu}$. Append the  tree of depth $d$ associated with the version space $V_{t-1}(y_{\mu}, \varepsilon_{\mu})$ to the edge indexed by $\mu$.  Note that the depth of $\mathcal{T}$ must be $d+1$. Furthermore, observe that for every hypothesis $h \in V_{t-1}(y_{\mu}, \varepsilon_{\mu})$, we have that $\mathbb{E}_{z \sim \mu}\left[\ell(y_{\mu}, z) \right] \geq \ell(y_{\mu}, h(x_t)) + \gamma$, matching the shattering condition in Definition \ref{def:SMdim}. Therefore, by definition of SMdim, we have that $\texttt{SM}_{\gamma}(V_{t-1}) \geq d+1$, a contradiction.  Thus, it must be the case that $\texttt{SM}_{\gamma}(V_t) \leq \texttt{SM}_{\gamma}(V_{t-1}) - 1$. 

Now, suppose $t \in [T]$ is a round such that $\texttt{SM}_{\gamma}(V_{t-1}) = 0$. We show that there always exist a distribution $\mu_t \in \Pi(\mathcal{Z})$ such that for all $(y, \varepsilon) \in \mathcal{C}_t$ , we have $\mathbb{E}_{z_t \sim \mu_t}\left[\ell(y, z_t) \right] < \gamma + \varepsilon$. Since we are in the $\varepsilon_t$-realizable setting, it must be the case that $(y_t, \varepsilon_t) \in \mathcal{C}_t$.  To see why such a $\mu_t$ must exist, suppose for the sake of contradiction that it does not exist. Then, for all $\mu \in \Pi(\mathcal{Z})$, there exists a pair $(y_{\mu}, \varepsilon_{\mu}) \in \mathcal{C}_t$ such that $\mathbb{E}_{z \sim \mu}\left[\ell(y_{\mu}, z)\right]  \geq \gamma + \varepsilon_{\mu} $. As before, consider a tree with root node labeled by $x_t$. For each measure $\mu \in \Pi(\mathcal{Z})$, construct an outgoing edge from $x_t$ indexed by $\mu$ and labeled by $y_{\mu}$. Since $(y_{\mu}, \varepsilon_{\mu}) \in \mathcal{C}_t$, there exists a hypothesis $h_{\mu} \in V_{t-1}$ such that $\ell(y_{\mu}, h_{\mu}(x_t)) \leq \varepsilon_{\mu}$. Therefore, we have $\mathbb{E}_{z \sim \mu}\left[\ell(y_{\mu}, z) \right] \geq \ell(y_{\mu}, h_{\mu}(x_t)) + \gamma$ . By definition of SMdim, this implies that $\texttt{SM}_{\gamma}(V_{t-1}) \geq 1$, which contradicts the fact that $\texttt{SM}_{\gamma}(V_{t-1}) = 0$. Thus, there must be a distribution $\mu_t \in \Pi(\mathcal{Z})$ such that for for all $(y, \varepsilon) \in \mathcal{C}_t$ , we have $\mathbb{E}_{z \sim \mu_t}\left[\ell(y, z) \right] < \gamma + \varepsilon$. Since this is precisely the distribution that Algorithm \ref{alg:rand_SOA} plays whenever $\texttt{SM}_{\gamma}(V_{t-1}) = 0$  and since $\texttt{SM}_{\gamma}(V_{t'}) \leq \texttt{SM}_{\gamma}(V_{t-1})$ for all $t' \geq t$, the algorithm no longer suffers expected loss more than $\gamma + \varepsilon_{t'}$ for all $t^{\prime} \geq t$. This completes the proof. 

\section{Proof of Lowerbound in Theorem \ref{thm:agn}}\label{proof:agn_lower}
    We now prove the lowerbound in Theorem \ref{thm:agn}.  Fix $\gamma > 0$ and $d_{\gamma} := \texttt{SM}_{\gamma}(\mathcal{H}) $. By definition of SMdim, there exists a $\mathcal{X}$-valued, $\Pi(\mathcal{Z})$-ary tree $\mathcal{T}$ of depth $d_{\gamma}$ shattered by $\mathcal{H}$. Let $(f_1, ..., f_d)$ be the sequence of edge-labeling functions $f_t: \Pi(\mathcal{Z})^t \rightarrow \Ycal$ associated with $\mathcal{T}$. Let $\mathcal{A}$ be any randomized learner for $\mathcal{H}$. Our goal will be to use $\mathcal{T}$ and its edge-labeling functions $(f_1, ..., f_d)$ to construct a difficult stream for $\mathcal{A}$ such that on every round, the expected loss of $\mathcal{A}$ is at least $ \gamma$ more than the loss of the optimal hypothesis in hindsight. This stream is obtained by traversing $\mathcal{T}$ adapting to the sequence of distributions output by $\mathcal{A}$. 

To that end, for every round $t \in [d_{\gamma}]$,  let $\mu_t$ denote the distribution that $\mathcal{A}$ computes before making its prediction $z_t \sim \mu_t$. Consider the stream $\{\left(\mathcal{T}_t(\mu_{<t}), f_t(\mu_{\leq t})\right)\}_{t=1}^{d_{\gamma}}$, where $\mu = 
(\mu_1, \ldots, \mu_{d_{\gamma}})$ denotes the sequence of distributions output by $\Acal$. This stream is obtained by starting at the root of $\mathcal{T}$, passing $\mathcal{T}_1$ to $\mathcal{A}$, observing the distribution $\mu_1$ computed by $\mathcal{A}$, passing the label $f_t(\mu_{\leq 1})$ to $\mathcal{A}$, and then finally moving along the edge indexed by $\mu_1$. This process then repeats $d_{\gamma} -1$ times until the end of the tree $\mathcal{T}$ is reached. Note that we can observe and use the distribution computed by $\mathcal{A}$ on round $t$ to generate the label because $\mathcal{A}$ \textit{deterministically} maps a sequence of labeled instances to a distribution.

Recall that the shattering condition implies that $\exists h_{\mu} \in \mathcal{H}$ such that $ \expect_{z_t \sim \mu_t}[\ell(f_t(\mu_{\leq t}), z_t)] \geq \ell(f_t(\mu_{\leq t}), h_{\mu}(\mathcal{T}_t(\mu_{<t})) + \gamma $ for all $t \in [d_{\gamma}]$. Therefore, the regret of $\Acal$ on the stream described above is at least 
\begin{equation*}
    \begin{split}
      \texttt{R}_{\Acal}(T, \mathcal{H}, \ell) \geq  \sum_{t=1}^{d_{\gamma}} \expect_{z_t \sim \mu_t}[\ell(f_t(\mu_{\leq t}), z_t)]  -  \sum_{t=1}^{d_{\gamma}}   \ell(f_t(\mu_{\leq t}), h_{\mu}(\mathcal{T}_t(\mu_{<t})) \geq \sum_{t=1}^{d_{\gamma}}\gamma = \gamma d_{\gamma}.
    \end{split}
\end{equation*}
 Since our choice of $\gamma$ and the randomized algorithm $\mathcal{A}$ is arbitrary, this holds true for any $\gamma > 0$ and randomized online learner. This completes our proof.

\section{Proof of Lemma \ref{lem:rel_SMdim_pdim}} \label{app:rel_SMdim_pdim}
 Let  $p = \texttt{H}(\mathcal{Y}, \mathcal{Z}, \ell)$. Fix $\gamma \in (0, 1]$ and $\gamma' < \gamma$.  Let $\mathcal{T}$ be a $\mathcal{X}$-valued, $\Pi(\mathcal{Z})$-ary tree of depth $d = \texttt{SM}_{\gamma}(\mathcal{H})$ shattered by $\mathcal{H}$. Let $v$ be the root node of $\mathcal{T}$ and $x$ denote the instance labeling $v$. Recall that $v$ has an outgoing edge for each measure $\mu \in \Pi(\mathcal{Z})$. In particular, this means that $v$ has outgoing edges corresponding to the Dirac measures on $\mathcal{Z}$, which we denote by $\{\delta_z\}_{z \in \mathcal{Z}}$. Fix a $z \in \mathcal{Z}$ and consider the outgoing edge from $v$ indexed by $\delta_z$. Let $y_z \in \mathcal{Y}$ be the element that labels the outgoing edge indexed by $\delta_z$. Let $\mathcal{H}_{\delta_z} = \{h_{\mu}: \mu \in \Pi(\mathcal{Z})^{d}, \mu_1 = \delta_z\} \subset \mathcal{H}$ denote the set of shattering hypothesis that corresponds to following a path down $\mathcal{T}$ that takes the edge $\delta_z$ in the first level. Then, for all $h \in \mathcal{H}_{\delta}$, the shattering condition implies that

$$\ell(y_z, z) \geq \ell(y_z, h(x)) + \gamma > \ell(y_z, h(x)) + \gamma'.$$

Taking the supremum on both sides further gives that

$$\ell(y_z, z) > \sup_{h \in \mathcal{H}_{\delta_z}}\ell(y_z, h(x)) + \gamma'.$$

Let $B_z := \texttt{B}_{\ell}(y_z, r_z + \gamma') \in \texttt{B}_{\ell}(\mathcal{Y})$ be the ball centered around $y_z$ of radius $r_z + \gamma^{\prime}$ where $r_z := \sup_{h \in \mathcal{H}_{\delta_z}}\ell(y_z, h(x))$. The inequality above implies that $z \notin B_z$ (note that $B_z$ changes depending on $z$). Since $z \in \mathcal{Z}$ was arbitrary, it must be the case that $z \notin B_z$ for all $z \in \mathcal{Z}$. This means that $\bigcap_{z \in \mathcal{Z}}B_z = \emptyset$. Then, using the fact that $(\mathcal{Y}, \mathcal{Z}, \ell)$ is a Helly space with Helly number $p$, there exists $p$ balls in $\{B_z\}_{z \in \mathcal{Z}}$ such that their collection-wise intersection is also empty. Accordingly, let $z_1, ..., z_p$ be such that $\bigcap_{i=1}^p B_{z_i} = \emptyset$. As before, for every $i \in [p]$, let $y_{z_i}$ denote the label on the outgoing edge from $v$ indexed by the Dirac measure $\delta_{z_i}$. By definition, for all $i \in [p]$ and $h_{\delta_{z_i}} \in \mathcal{H}_{\delta_{z_i}}$ we have that 

$$h_{\delta_{z_i}}(x) \in \texttt{B}_{\ell}(y_{z_i}, r_{z_i}) := \tilde{B}_{z_i}$$

Note that $B_{z_i}$ is the $\gamma'$ expansion of  $\tilde{B}_{z_i}$. For each $i \in [p]$, relabel the outgoing edge from $v$ indexed by $\delta_{z_i}$ with the tuple $(y_{z_i},r_{z_i})$. For each $i \in [p]$, reindex the outgoing edge from $v$ indexed by $\delta_{z_i}$ with $i$. Remove all other edges indexed by measures and their corresponding subtrees. There should now only be $p$ outgoing edges from $v$, each indexed by a number $i \in [p]$ and labeled by a tuple in $\mathcal{Y} \times [0, c]$. Note that  $\bigcap_{i=1}^p \texttt{B}_{\ell}(y_{z_i}, r_{z_{i}} + \gamma') = \bigcap_{i=1}^p B_{z_i} = \emptyset$, which matches the second constraint imposed by Definition \ref{def:pdim}. As for the first constraint on shattering, note that for all $i \in [p]$ and all $h_{\delta_{z_i}} \in \mathcal{H}$,  we have that $h_{\delta_{z_i}}(x) \in \tilde{B}_{z_i}$. Thus, the hypothesis that shatters the edges indexed by $\delta_{z_i}$ in the original tree according to Definition \ref{def:SMdim} also shatters the newly re-indexed and relabeled edges according to Definition \ref{def:pdim}. Thus, for the root node $v$, both constraints imposed by Definition \ref{def:pdim} are met. Recursively repeating the above procedure on the subtrees following the $p$ newly reindexed and relabeled edges results in a $p$-dim tree $\gamma'$-shattered by $\mathcal{H}$ of depth $d$. Thus, $\texttt{SM}_{\gamma}(\mathcal{H}) \leq p\texttt{-dim}_{\gamma'}(\mathcal{H})$ for $\gamma' < \gamma$.

\end{document}